\documentclass[10pt,twocolumn,letterpaper]{article}

\usepackage[pagenumbers]{cvpr}   %

\DeclareMathOperator*{\argmax}{arg\,max}

\usepackage{tabularx}
\usepackage{colortbl}
\usepackage[normalem]{ulem}
\usepackage{pifont}
\usepackage{fontawesome5}

\definecolor{codewampurple}{HTML}{7B2CBF}
\definecolor{tabmain}{HTML}{EEF2F8}
\definecolor{tabours}{HTML}{E8F0F8}
\definecolor{tabgroup}{HTML}{F2F4F7}
\definecolor{tabheadbg}{HTML}{F7F9FC}
\definecolor{tabink}{HTML}{34465B}
\definecolor{tabrule}{HTML}{7B8B9F}
\definecolor{tabyescolor}{RGB}{34,139,34}
\definecolor{tabnocolor}{RGB}{200,40,40}
\definecolor{tabbestcolor}{HTML}{B2182B}
\definecolor{tabsecondcolor}{HTML}{1F4E9C}
\definecolor{tabthirdcolor}{HTML}{1B7A6B}

\newcommand{\tabnum}[1]{#1}
\newcommand{\tabvenue}[1]{{\color{tabink}#1}}
\newcommand{\tabhead}[1]{\textcolor{tabink}{\textbf{#1}}}
\newcommand{\tabgrouplabel}[1]{{\color{tabink}\bfseries #1}}
\newcommand{\tabyes}{{\color{tabyescolor}\ding{51}}}
\newcommand{\tabno}{{\color{tabnocolor}\ding{55}}}

\newcommand{\tabrank}[1]{\textsuperscript{\normalfont\itshape\fontsize{4.5}{5}\selectfont #1}}
\newcommand{\tabbest}[1]{\textcolor{tabbestcolor}{\mbox{\textbf{#1}\tabrank{1st}}}}
\newcommand{\tabsecond}[1]{\textcolor{tabsecondcolor}{\mbox{#1\tabrank{2nd}}}}
\newcommand{\tabthird}[1]{\textcolor{tabthirdcolor}{\mbox{#1\tabrank{3rd}}}}
\newcommand{\tabbestkey}{\textcolor{tabbestcolor}{\textbf{1st}}}
\newcommand{\tabsecondkey}{\textcolor{tabsecondcolor}{2nd}}
\newcommand{\tabthirdkey}{\textcolor{tabthirdcolor}{3rd}}

\newcommand{\tabstyle}{%
  \renewcommand{\arraystretch}{1.18}%
  \setlength{\aboverulesep}{0.35ex}%
  \setlength{\belowrulesep}{0.45ex}%
  \arrayrulecolor{tabrule}%
}

\newcommand{\tabstages}[2]{\begin{tabular}{@{}c@{}}#1\\#2\end{tabular}}

\definecolor{cvprblue}{rgb}{0.21,0.49,0.74}
\usepackage[pagebackref,breaklinks,colorlinks,allcolors=cvprblue]{hyperref}

\def\confName{CVPR}
\def\confYear{2027}

\title{\textcolor{codewampurple}{\uuline{\textit{CodeWAM}}}: A Code World Action Model for Autonomous Driving}

\author{%
Zhisheng Chen$^{1,*}$ \quad Jinpeng Wang$^{2,*}$ \quad
Beicheng Lou$^{3}$ \quad Zhengwei Xie$^{4}$\\
Hao Wu$^{2}$ \quad Rui Qian$^{5,\dagger}$ \quad
Kun Wang$^{1,\dagger}$ \quad Vivian W. H. Wong$^{6}$\\[0.4em]
{\small $^{1}$Nanyang Technological University
\quad $^{2}$Tsinghua University
\quad $^{3}$Independent Researcher}\\
{\small $^{4}$University of Science and Technology of China
\quad $^{5}$Fudan University
\quad $^{6}$University of Florida}\\[0.3em]
{\small $^{*}$Equal contribution. \quad $^{\dagger}$Corresponding authors.}\\[0.2em]
{\small \faGithub\enspace \href{https://github.com/Zhisheng-researcher/CodeWAM}{\texttt{github.com/Zhisheng-researcher/CodeWAM}}
\quad \faEnvelope\enspace \href{mailto:zhisheng.researcher@gmail.com}{\texttt{zhisheng.researcher@gmail.com}}}%
}

\begin{document}
\maketitle
\begin{abstract}
Driving world models anticipate action consequences through representations of how the world evolves. For autonomous driving, these representations must capture both the surrounding scene and the ego vehicle's motion under a candidate plan. We propose CodeWAM, a driving world model that incorporates executable code as part of its world representation. Code represents ego dynamics by expressing how the vehicle state evolves under plan execution, while a learned visual model captures scene-dependent consequences. Executing the code produces a candidate-conditioned motion sequence, which is evaluated relative to the prior plan and recorded motion history. The resulting execution consequences and learned scene outcomes jointly represent the candidate's future, enabling the planner to assess both what a plan may lead to and how the vehicle transitions into that future. The selected plan becomes part of the next decision's execution context, linking successive choices within the model. CodeWAM achieves 93.6 PDMS on NAVSIM v1 navtest, 92.1 EPDMS on v2 navtest, and 50.6 EPDMS on NAVHARD. Further experiments demonstrate improved pairwise execution consistency and gains on Hydra-MDP and DrivoR without additional training, supporting executable code as a complementary representation for driving world models.
\end{abstract}

\section{Introduction}
\label{sec:intro}

World models enable autonomous driving systems to predict the consequences of actions before taking them and select better plans accordingly.
Recent work compares driving options by predicting future images, scene features, or the reward of a candidate trajectory, and selects the safer and more effective behavior \cite{zeng2026futuresightdrive,zheng2025world4drive,zhang2026future}.
These methods rely mainly on neural networks to learn from driving data how the world changes.
Code world models offer another route: the change of the vehicle state under the execution of a plan is written as a program, and running the program rolls out the motion that follows if a candidate is adopted \cite{li2025generalized,tang2024worldcoder}.
The execution process described by code thus becomes part of the world model, and the predicted ego motion, together with the predicted surroundings, forms the judgment of a candidate's consequence.

\begin{figure}[!t]
\centering
\includegraphics[width=\columnwidth]{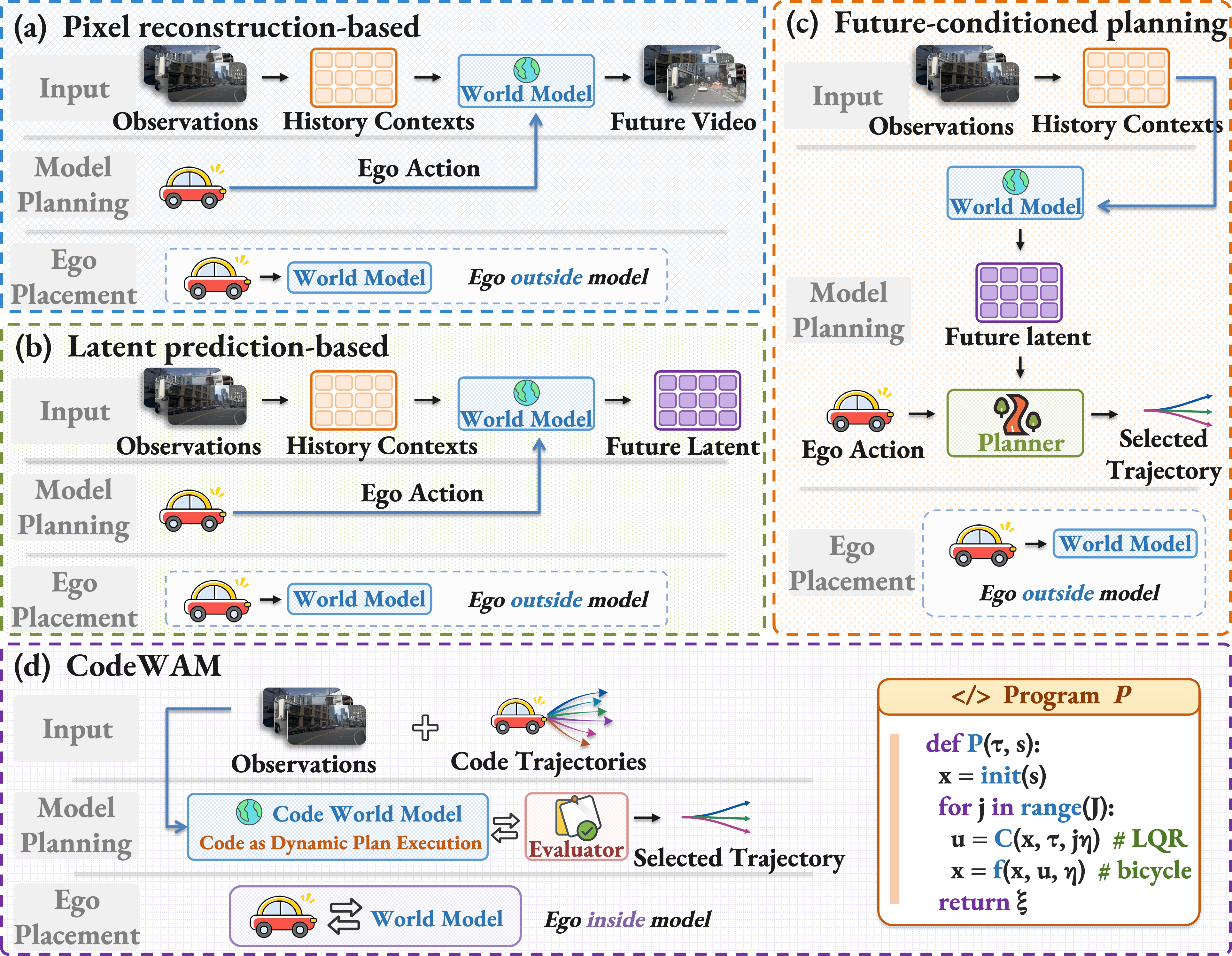}
\caption{Comparison of driving world model paradigms. (a–c) Predictive representations support planning through future video, latent prediction, or future conditioning. (d) CodeWAM explicitly models candidate execution from current states with executable ego dynamics, combining the resulting motion consequences with learned scene outcomes for trajectory selection.}
\label{fig:comparison}
\end{figure}

How this execution process is represented directly shapes what a world model judges about the value of a candidate plan.
Whenever the vehicle makes a new decision, it is already in a motion state formed by its previous actions, and the new plan must continue from that state.
For example, a rightward correction may look safe and efficient on its own, but if the vehicle is turning left, adopting it at once may cause an unnecessary sway.
The value of a candidate therefore depends not only on the future it points to, but also on how the vehicle moves from its current motion into that future.
If this process is ignored, a world model may overrate candidates that have a high expected reward but cannot be executed smoothly.
The planner then departs from the reward the vehicle can actually realize, and successive decisions produce conflicting motion adjustments.

We therefore propose CodeWAM (Code World Action Model), which uses executable code as a representation of the world model and predicts the consequence of a candidate plan together with a learned model.
Figure~\ref{fig:comparison} shows the difference.
CodeWAM describes with a program how the ego vehicle executes a new plan: given the current vehicle state and the execution information left by the previous plan, it computes, under a given vehicle motion model, the motion that each candidate would produce if adopted.
The scene world model takes multi-view images as observation and predicts the consequence a candidate may have in the surroundings, including collision risk and road rules.
The planner combines the two predictions to select a plan, and at the next decision it again takes into account the vehicle has already produced.
The link between successive plans becomes something the world model describes explicitly.

In CodeWAM, code carries the representation of ego dynamics: how the current motion continues, and how a new plan changes that motion, are both described by an executable program.
Executing this code gives the predicted ego motion for each candidate, so the process from proposing a plan to executing it enters the world model's description of the future.
For the external scene, prediction remains uncertain, and overrated candidates are more likely to win the selection.
We therefore train the scene world model on the states the planner itself will meet, and constrain candidate selection with the prediction error obtained by calibration.
The two representations together describe the future of a candidate plan: the code describes how the ego vehicle evolves from its current motion to the execution of a new plan, and the scene predicts what this behavior may cause in the surroundings.
Both provide the basis for the decision.

We evaluate CodeWAM on navtest and navhard of NAVSIM \cite{dauner2024navsim}, and study zero-shot transfer to nuScenes \cite{caesar2020nuscenes} after training on NAVSIM only.
The experiments consider driving performance and whether successive plans connect smoothly, and examine each design by removing the program and by removing the information of the previous plan.
We also study how behavior changes as the number of candidates grows, and connect the program to different methods to test whether it applies without additional training.

Our contributions are threefold:
\begin{itemize}
    \item \textbf{An executable representation of ego dynamics for driving world models.}
    We represent plan execution with code that computes ego motion under candidate plans. Combined with learned scene consequences, this representation connects the future a plan entails with the vehicle motion required to realize it.

    \item \textbf{CodeWAM for execution-aware continuous planning.}
    CodeWAM evaluates candidate execution relative to the preceding plan and recorded motion history, jointly with predicted scene consequences. The selected plan becomes part of the next decision's context, explicitly linking successive choices through their predicted motion.

    \item \textbf{Empirical validation across planning methods.}
    Experiments show that the executable ego model improves multiple planners without additional training, demonstrating the value of incorporating execution consequences into trajectory evaluation across different architectures.
\end{itemize}

\section{Related Work}
\label{sec:related}

\begin{figure*}[t]
\centering
\includegraphics[width=\textwidth]{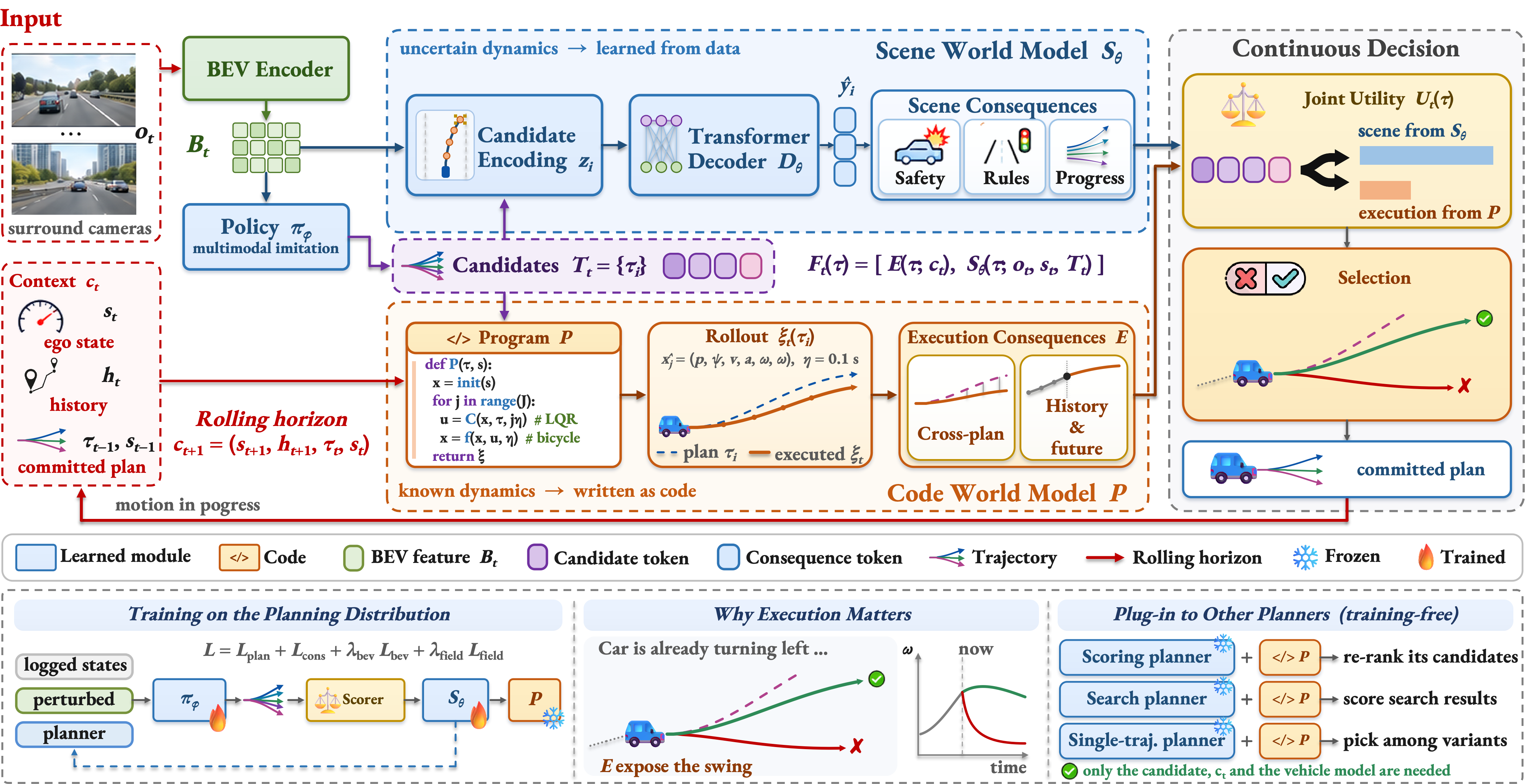}
\caption{Overview of CodeWAM. For each candidate plan proposed by the policy, a program rolls out the motion the ego vehicle produces from its current state, and the resulting motion is compared with the previous plan and with the recorded history to obtain the execution consequence. A learned scene model predicts, from multi-view images, the consequence of the same candidate in the scene. The two enter a joint utility, and the plan selected at this step becomes the execution context of the next one.}
\label{fig:overview}
\end{figure*}

\paragraph{Driving world models and trajectory evaluation.}
Driving world models anticipate action consequences to support planning.
Drive-WM \cite{wang2024driving} generates action-conditioned multi-view videos for reward-based planning, while WoTE \cite{li2025end} predicts future BEV states to evaluate trajectory candidates.
Latent-WAM \cite{wang2026latent} jointly models scene and ego status in latent space, and DriveFuture \cite{hong2026drivefuture} conditions trajectory generation on predicted future representations.
Complementing these predictive approaches, GTRS \cite{li2025generalized} improves scoring across diverse candidate distributions, and DrivoR \cite{kirby2026driving} learns interpretable driving sub-scores from compact visual features.
Together, these studies establish future prediction and candidate evaluation as effective mechanisms for driving decisions.
CodeWAM incorporates plan execution into world model consequence prediction, combining executable ego dynamics with learned scene consequences in a joint representation. This enables candidate evaluation to account for both the future a plan entails and how the vehicle transitions into that future from its current state.

\paragraph{Executable programs for embodied world modeling.}
Programs provide a structured representation that can be executed for control and prediction.
Code as Policies \cite{opdebeeck2026empirical} generates policy programs that compose perception and control APIs.
WorldCoder \cite{tang2024worldcoder} and GIF-MCTS \cite{dainese2024generating} construct code world models through program synthesis and feedback, enabling planning with executable environment models.
AW-0 \cite{liu2026agentic} jointly generates skill programs, future video, and robot actions, using code to organize prediction and support reusable skill execution.
CodeWAM uses code as an executable representation of ego dynamics: a specified controller and vehicle model map candidate plans to motion sequences. This computation complements visual consequence prediction within a shared representation.

\paragraph{Execution modeling and temporal planning.}
Model-based planning and temporal planning provide closely related foundations for evaluating how decisions unfold.
PDM-Closed \cite{dauner2023parting} simulates candidate execution with an LQR controller and a kinematic bicycle model;  PDM-Hybrid \cite{dauner2023parting} further introduces learned corrections to longer-horizon waypoints.
NAVSIM v2 \cite{cao2025pseudo} uses controller-based simulation for evaluation and provides history and cross-plan comfort criteria.
Temporal planning methods incorporate past decisions into current planning: MomAD \cite{song2025don} combines prior-path matching with historical planning queries, while TemporalDiffuser \cite{zhang2026temporaldiffuser} conditions trajectory generation on historical planning states.
CodeWAM evaluates each candidate's predicted motion against the preceding plan and recorded history, combining temporal execution consequences with learned scene outcomes. The selected plan becomes the next decision's execution context, linking successive choices through their predicted motion.

\section{Method}
\label{sec:method}

\subsection{Overall Framework}
\label{sec:method-overall}

CodeWAM predicts the consequence of a candidate plan in two parts (Figure~\ref{fig:overview}): a program, which derives the motion the ego vehicle produces when it executes the plan, and a learned scene model, which predicts the reaction of the surrounding scene.
Whether another vehicle yields or a pedestrian crosses is decided by intent and interaction and can only be summarized from data, whereas how the ego vehicle moves under a new plan follows from its current state and can be derived once the vehicle model is given.

At decision step $t$, the input consists of multi-view images $o_t$, the vehicle state $s_t$, the motion history $h_t$, the plan $\tau_{t-1}$ selected at the previous step, and the state $s_{t-1}$ at which it was selected.
Equations the execution context
\begin{equation}
c_t=(s_t,h_t,\tau_{t-1},s_{t-1}).
\label{eq:context}
\end{equation}
The policy $\pi_\phi$ generates a candidate set $\mathcal T_t$ and takes the mode with the highest confidence as the default plan $\tau_t^0$; the set also contains plans obtained by continuing the previous plan.
Each candidate $\tau=\{(p_k,\psi_k)\}_{k=1}^{K}$ gives positions and headings over $H$ seconds at a waypoint interval $\Delta=H/K$. $\delta_t$ denotes decision interval and $\eta$ the rollout step.

For each candidate, CodeWAM predicts
\begin{equation}
\begin{aligned}
\xi_t(\tau)&=\mathcal P(\tau;s_t),\\
e_t(\tau)&=\mathcal R(\xi_t(\tau),c_t),\\
\hat y_t(\tau)&=\mathcal S_\theta(\tau;o_t,s_t,\mathcal T_t),\\
\mathcal F_t(\tau;\mathcal T_t)&=[e_t(\tau),\hat y_t(\tau)].
\end{aligned}
\label{eq:joint}
\end{equation}
$\mathcal R$ computes the execution consequence from the execution motion and the execution context.
The scene model attends across candidates, so $\hat y_t$ depends on the whole set.
The planner selects a plan from $\mathcal F_t$.

\eqref{eq:joint} also locates a common learned candidate scorer, which predicts a consequence directly from waypoints and scene, i.e.\ keeps only $\hat y_t$.
Such a scorer has to learn the execution process from data, so its output depends on whether the labels cover the execution states the vehicle meets.
CodeWAM computes $e_t$ with a program, and its evaluation depends on the current motion and previous decision.

\subsection{Program Representation of Ego Dynamics}
\label{sec:method-program}

A candidate gives the positions the vehicle is expected to pass; how it actually moves also depends on its current speed, acceleration, and yaw.
We represent this with an executable program and roll it out step by step:
\begin{equation}
\begin{aligned}
\xi_t(\tau)&=\{\hat x_j\}_{j=0}^{J},\quad \hat x_j=(p_j,\psi_j,v_j,a_j,\omega_j,\dot\omega_j),\\
u_j&=C(\hat x_j,\tau,j\eta),\quad
\hat x_{j+1}=f(\hat x_j,u_j;\eta),
\end{aligned}
\label{eq:transition}
\end{equation}
where $J=H/\eta$, $\omega_j$ and $\dot\omega_j$ are the yaw rate and yaw acceleration, and a tracking controller $C$ produces the input $u_j$ that a state transition $f$ turns into the next state.
The rollout starts from $s_t$, with positions in the current ego frame.

Because the controller reads the current state at every step, the execution motion differs from the candidate.
Speeds and accelerations obtained by differentiating waypoints are not constrained by the current motion, whereas $\xi_t(\tau)$ starts from $s_t$ and carries the tracking error, the change of acceleration, and the change of yaw that the transition produces.
If the vehicle is turning left, a rightward candidate can be tracked only after the yaw rate has fallen back, and this appears only in the execution motion.

\paragraph{Execution consequence.}
When the vehicle updates its plan, the previous plan is still under execution, so the consequence of a candidate is measured against two motions.
For the previous plan we roll out $\xi_{t-1}=\mathcal P(\tau_{t-1};s_{t-1})$, align it to the current frame and time by $\delta_t$, and obtain $\bar\xi_{t-1}$; for each motion quantity $q\in\{a,\dot a,\omega,\dot\omega\}$ the difference over the overlapping samples $\mathcal J_t=\{j: j\eta\le H-\delta_t\}$ is
\begin{equation}
d_{t,q}(\tau)=
\left[
\frac{1}{|\mathcal J_t|}
\sum_{j\in\mathcal J_t}
\left\|q\big(\xi_t(\tau)_j\big)
-q\big(\bar\xi_{t-1,j}\big)\right\|_2^2
\right]^{1/2},
\label{eq:diff}
\end{equation}
with $\dot a$ the jerk.
For the recorded history we interpolate $h_t$ onto the rollout grid and concatenate it with the execution motion, which yields a sequence $\zeta_t(\tau)$ crossing the current time; its changes of acceleration, jerk, and yaw show whether the vehicle moves smoothly into the candidate.
The execution consequence is
\begin{equation}
e_t(\tau)=
\left[
\rho_{\mathrm{cross}}\big(d_t(\tau)\big),\;
\rho_{\mathrm{hist}}\big(\zeta_t(\tau)\big)
\right],
\label{eq:exec}
\end{equation}
with a higher value meaning a smoother link.
The two maps compare the execution motion with a reference motion and return a value in $[0,1]$.
There are two references, the motion predicted by the previous plan and the recorded motion history.
A higher value means the two are closer and the link is smoother.
The previous plan and the history act on the evaluation only through \eqref{eq:exec} and do not change the rollout.

\subsection{Prediction of Scene Consequence}
\label{sec:method-scene}

The scene model encodes the four surround-view images with DINOv2 adapted by LoRA.
Using the camera parameters, it projects 3D sample points of a BEV grid onto the image features, aggregates them over visible cameras, and passes them through a convolutional backbone to a scene feature $B_t$ shared with the policy.
Because the features are sampled geometrically, the same images can be resampled under another ego pose, which Section~\ref{sec:method-training} uses to build training states.
For a candidate $\tau_i$ we sample $B_t$ along its waypoints and at the four corners of the vehicle footprint, and sample a predicted map distance field $D_t$ in the same places, giving the position of the candidate with respect to the drivable area and the lane centerline:
\begin{equation}
\begin{aligned}
z_i=\operatorname{Fuse}\big(&
\operatorname{Embed}(\tau_i),
\operatorname{Sample}(B_t,\tau_i),\\
&\operatorname{Sample}_{\mathrm{corner}}(B_t,\tau_i),
\operatorname{Sample}_{\mathrm{map}}(D_t,\tau_i)\big).
\end{aligned}
\label{eq:candidate}
\end{equation}
All candidates enter a Transformer decoder that attends across candidates and to $B_t$:
\begin{equation}
(\hat y_1,\ldots,\hat y_N)
=\mathcal D_\theta\big((z_1,\ldots,z_N),B_t,s_t\big),
\qquad
\hat y_i\in[0,1]^{d_c}.
\label{eq:decoder}
\end{equation}
$\hat y_i$ covers safety risk, road compliance, and driving quality, supervised on $[0,1]$.
The execution and scene consequences are obtained differently but refer to the same candidate, so they combine directly in \eqref{eq:joint}; they differ in training, since the program rolls out from any state while the scene model must be trained on the states the planner meets.

\begin{table*}[!tbp]
\centering
\caption{Comparison on NAVSIM v1 navtest. All metrics are percentages ($\uparrow$). C: camera; L: LiDAR; MV-C: multi-view cameras; Ego: ego-state input. Colors and superscripts mark the \tabbestkey{}, \tabsecondkey{}, and \tabthirdkey{} distinct values per metric across all methods; ties share a rank.}
\label{tab:navsim-v1}
\footnotesize
\setlength{\tabcolsep}{2.0pt}
\tabstyle
\begin{tabularx}{\textwidth}{l<{\hspace{4.0pt}} l<{\hspace{4.0pt}} l<{\hspace{2.0pt}} c<{\hspace{0.0pt}} *{5}{>{\centering\arraybackslash}X} >{\bfseries\centering\arraybackslash}X}
\toprule
\multicolumn{4}{c}{\textbf{}} & \multicolumn{6}{c}{\textbf{NAVSIM v1 metrics (\%) $\uparrow$}} \\
\cmidrule(lr){5-10}
\rowcolor{tabheadbg}\textbf{Method} & \textbf{Venue} & \textbf{Sensors} & \textbf{Ego} & \tabhead{NC} & \tabhead{DAC} & \tabhead{EP} & \tabhead{TTC} & \tabhead{Comfort} & \tabhead{PDMS} \\
\midrule
\rowcolor{tabgroup}\multicolumn{10}{l}{\tabgrouplabel{End-to-End Methods}} \\
UniAD \cite{hu2023planning} & \tabvenue{CVPR'23} & 6$\times$C & \tabyes & \tabnum{97.8} & \tabnum{91.9} & \tabnum{78.8} & \tabnum{92.9} & \tabbest{100.0} & \tabnum{83.4} \\
TransFuser \cite{chitta2022transfuser} & \tabvenue{TPAMI'23} & 3$\times$C+L & \tabyes & \tabnum{97.7} & \tabnum{92.8} & \tabnum{79.2} & \tabnum{92.8} & \tabbest{100.0} & \tabnum{84.0} \\
iPad \cite{guo2026ipad} & \tabvenue{RA-L'26} & 4$\times$C & \tabyes & \tabnum{98.6} & \tabnum{98.3} & \tabsecond{88.0} & \tabnum{94.9} & \tabbest{100.0} & \tabsecond{91.7} \\
WoTE \cite{li2025end} & \tabvenue{ICCV'25} & 3$\times$C+L & \tabyes & \tabnum{98.5} & \tabnum{96.8} & \tabnum{81.9} & \tabnum{94.9} & \tabsecond{99.9} & \tabnum{88.3} \\
DiffusionDrive \cite{liao2025diffusiondrive} & \tabvenue{CVPR'25} & 3$\times$C+L & \tabyes & \tabnum{98.2} & \tabnum{96.2} & \tabnum{82.2} & \tabnum{94.7} & \tabbest{100.0} & \tabnum{88.1} \\
SeerDrive \cite{zhang2026future} & \tabvenue{NeurIPS'25} & 3$\times$C+L & \tabyes & \tabnum{98.4} & \tabnum{97.0} & \tabnum{83.2} & \tabnum{94.9} & \tabsecond{99.9} & \tabnum{88.9} \\
WorldRFT \cite{yang2026worldrft} & \tabvenue{AAAI'26} & 3$\times$C & \tabyes & \tabnum{97.8} & \tabnum{96.8} & \tabnum{81.7} & \tabnum{94.0} & \tabbest{100.0} & \tabnum{87.8} \\
\rowcolor{tabgroup}\multicolumn{10}{l}{\tabgrouplabel{VLM-based Methods}} \\
UniVLA \cite{bu2025univla} & \tabvenue{ICLR'26} & 1$\times$C & \tabyes & \tabnum{96.9} & \tabnum{91.1} & \tabnum{76.8} & \tabnum{91.7} & \tabnum{96.7} & \tabnum{81.7} \\
AutoVLA \cite{zhou2026autovla} & \tabvenue{NeurIPS'25} & 3$\times$C & \tabyes & \tabnum{98.4} & \tabnum{95.6} & \tabnum{81.9} & \tabbest{98.0} & \tabsecond{99.9} & \tabnum{89.1} \\
FSDrive \cite{zeng2026futuresightdrive} & \tabvenue{NeurIPS'25} & MV-C & \tabno & \tabnum{98.2} & \tabnum{93.8} & \tabnum{80.1} & \tabnum{93.3} & \tabsecond{99.9} & \tabnum{85.1} \\
ReCogDrive \cite{xiong2026recogdrive} & \tabvenue{ICLR'26} & 1$\times$C & \tabyes & \tabnum{97.9} & \tabnum{97.3} & \tabthird{87.3} & \tabnum{94.9} & \tabbest{100.0} & \tabnum{90.8} \\
SGDrive \cite{li2026sgdrive} & \tabvenue{CVPR'26} & 1$\times$C & \tabyes & \tabnum{98.6} & \tabnum{97.8} & \tabnum{85.8} & \tabnum{96.2} & \tabbest{100.0} & \tabnum{91.1} \\
ExploreVLA \cite{sheng2026explorevla} & \tabvenue{ECCV'26} & 1$\times$C & \tabyes & \tabthird{98.8} & \tabthird{98.4} & \tabnum{83.5} & \tabthird{96.5} & \tabsecond{99.9} & \tabnum{90.4} \\
DriveVLA-W0 \cite{li2026drivevla} & \tabvenue{ICLR'26} & 1$\times$C & \tabyes & \tabnum{98.7} & \tabbest{99.1} & \tabnum{83.3} & \tabnum{95.3} & \tabthird{99.3} & \tabnum{90.2} \\
\rowcolor{tabgroup}\multicolumn{10}{l}{\tabgrouplabel{WM-based Methods}} \\
World4Drive \cite{zheng2025world4drive} & \tabvenue{ICCV'25} & 3$\times$C & \tabno & \tabnum{97.4} & \tabnum{94.3} & \tabnum{79.9} & \tabnum{92.8} & \tabbest{100.0} & \tabnum{85.1} \\
Epona \cite{zhang2025epona} & \tabvenue{ICCV'25} & 1$\times$C & \tabyes & \tabnum{97.9} & \tabnum{95.1} & \tabnum{80.4} & \tabnum{93.8} & \tabsecond{99.9} & \tabnum{86.2} \\
SimWAM \cite{zhao2026simwam} & \tabvenue{arXiv'26} & 1$\times$C & \tabyes & \tabnum{98.4} & \tabsecond{98.7} & \tabnum{86.4} & \tabnum{95.5} & \tabbest{100.0} & \tabthird{91.5} \\
DriveLaW \cite{xia2026drivelaw} & \tabvenue{CVPR'26} & 1$\times$C & \tabyes & \tabsecond{99.0} & \tabnum{97.1} & \tabnum{81.3} & \tabsecond{96.7} & \tabbest{100.0} & \tabnum{89.1} \\
\midrule
\rowcolor{tabours}\textbf{CodeWAM} {\scriptsize(Ours)} & & 4$\times$C & \tabyes & \tabbest{99.6} & \tabbest{99.1} & \tabbest{90.2} & \tabthird{96.5} & \tabbest{100.0} & \tabbest{93.6} \\
\bottomrule
\end{tabularx}
\end{table*}

\subsection{Decision with the Joint Consequence}
\label{sec:method-decision}

The planning utility is
\begin{equation}
U_t(\tau;\mathcal T_t)
=G\big(\mathcal F_t(\tau;\mathcal T_t)\big),
\label{eq:utility}
\end{equation}
where $G$ multiplies the safety and compliance components and averages driving quality with the execution consequence, following EPDMS (see supplementary material).
The scene consequence is learned and carries error, so selecting the highest utility directly would favor overrated candidates.
We therefore take the default plan $\tau_t^0$ as reference and replace it only when a candidate improves on its scene utility by $\lambda$:
\begin{equation}
\begin{aligned}
\mathcal A_t(\lambda)
&=\{\tau_t^0\}\cup
\left\{\tau\in\mathcal T_t:
\widehat U_t(\tau;\mathcal T_t)
-\widehat U_t(\tau_t^0;\mathcal T_t)>\lambda\right\},\\
\tau_t^{\mathrm{sel}}
&=\underset{\tau\in\mathcal A_t(\lambda)}{\argmax}
\;U_t(\tau;\mathcal T_t).
\end{aligned}
\label{eq:selection}
\end{equation}
The gain $\lambda$ accounts for the error of the learned part; the execution consequence is computed, so it does not enter this comparison.
$\lambda$ is chosen on held-out data.

The policy candidates do not always contain the smoother options, so we also expand the set around the selected plan by a local search in the acceleration and yaw rate space, and by interpolations with the continuation of the previous plan and speed-scaled variants.
A result of the local search is adopted when it improves the utility by more than $\lambda_c$, in the same way, and does not worsen the link to the previous plan.
New candidates are rolled out and scored by the same $G$.
After $\tau_t$ is executed, the context becomes
\begin{equation}
c_{t+1}=(s_{t+1},h_{t+1},\tau_t,s_t),
\label{eq:update}
\end{equation}
so the next step rolls out the previous plan from its recorded start and compares it with the new candidates; shifting the control sequence of $\tau_t$ forward also yields the continuation candidate.
Each selection affects both the candidate set of the next decision and the way its candidates are evaluated.

\begin{table*}[!tbp]
\centering
\caption{Comparison on NAVSIM v2 navtest under the corrected, single-stage non-reactive protocol. EPDMS and its nine components are percentages ($\uparrow$). Colors and superscripts mark the \tabbestkey{}, \tabsecondkey{}, and \tabthirdkey{} distinct values per metric across all methods; ties share a rank.}
\label{tab:navsim-v2}
\footnotesize
\setlength{\tabcolsep}{0.5pt}
\tabstyle
\begin{tabularx}{\textwidth}{l<{\hspace{7.0pt}} l<{\hspace{7.0pt}} l<{\hspace{5.0pt}} c<{\hspace{3.0pt}} *{9}{>{\centering\arraybackslash}X} >{\bfseries}c}
\toprule
\multicolumn{4}{c}{\textbf{}} & \multicolumn{10}{c}{\textbf{NAVSIM v2 metrics (\%) $\uparrow$}} \\
\cmidrule(lr){5-14}
\rowcolor{tabheadbg}\textbf{Method} & \textbf{Venue} & \textbf{Sensors} & \textbf{Ego} & \tabhead{NC} & \tabhead{DAC} & \tabhead{DDC} & \tabhead{TLC} & \tabhead{EP} & \tabhead{TTC} & \tabhead{LK} & \tabhead{HC} & \tabhead{EC} & \textbf{EPDMS} \\
\midrule
\rowcolor{tabgroup}\multicolumn{14}{l}{\tabgrouplabel{End-to-End Methods}} \\
TransFuser \cite{chitta2022transfuser} & \tabvenue{TPAMI'23} & 3$\times$C+L & \tabyes & \tabnum{96.9} & \tabnum{89.9} & \tabnum{97.8} & \tabthird{99.7} & \tabnum{87.1} & \tabnum{95.4} & \tabnum{92.7} & \tabbest{98.3} & \tabnum{87.2} & \tabnum{76.7} \\
DriveSuprim \cite{yao2026drivesuprim} & \tabvenue{AAAI'26} & 3$\times$C & \tabno & \tabnum{97.5} & \tabnum{96.5} & \tabnum{99.4} & \tabnum{99.6} & \tabthird{88.4} & \tabnum{96.6} & \tabnum{95.5} & \tabbest{98.3} & \tabnum{77.0} & \tabnum{83.1} \\
DiffusionDrive \cite{liao2025diffusiondrive} & \tabvenue{CVPR'25} & 3$\times$C+L & \tabyes & \tabnum{98.2} & \tabnum{95.9} & \tabnum{99.4} & \tabsecond{99.8} & \tabnum{87.5} & \tabnum{97.3} & \tabnum{96.8} & \tabbest{98.3} & \tabsecond{87.7} & \tabnum{84.5} \\
Artemis \cite{feng2025artemis} & \tabvenue{RA-L'26} & 3$\times$C+L & \tabyes & \tabthird{98.3} & \tabnum{95.1} & \tabnum{98.6} & \tabsecond{99.8} & \tabnum{81.5} & \tabnum{97.4} & \tabnum{96.5} & \tabbest{98.3} & \tabnum{--} & \tabnum{83.1} \\
WorldRFT \cite{yang2026worldrft} & \tabvenue{AAAI'26} & 3$\times$C & \tabyes & \tabnum{97.8} & \tabnum{96.5} & \tabthird{99.5} & \tabsecond{99.8} & \tabsecond{88.5} & \tabnum{97.0} & \tabthird{97.4} & \tabsecond{98.1} & \tabnum{69.1} & \tabthird{86.7} \\
WoTE \cite{li2025end} & \tabvenue{ICCV'25} & 3$\times$C+L & \tabyes & \tabnum{97.4} & \tabnum{88.2} & \tabnum{97.8} & \tabnum{99.3} & \tabnum{82.7} & \tabnum{96.4} & \tabnum{90.9} & \tabnum{97.3} & \tabnum{68.0} & \tabnum{66.7} \\
\rowcolor{tabgroup}\multicolumn{14}{l}{\tabgrouplabel{VLM-based Methods}} \\
DriveVLA-W0 \cite{li2026drivevla} & \tabvenue{ICLR'26} & 1$\times$C & \tabyes & \tabsecond{98.5} & \tabbest{99.1} & \tabnum{98.0} & \tabthird{99.7} & \tabnum{86.4} & \tabsecond{98.1} & \tabnum{93.2} & \tabnum{97.9} & \tabnum{58.9} & \tabnum{86.1} \\
ReCogDrive \cite{xiong2026recogdrive} & \tabvenue{ICLR'26} & 1$\times$C & \tabyes & \tabthird{98.3} & \tabnum{95.2} & \tabthird{99.5} & \tabsecond{99.8} & \tabnum{87.1} & \tabthird{97.5} & \tabnum{96.6} & \tabbest{98.3} & \tabnum{86.5} & \tabnum{83.6} \\
\rowcolor{tabgroup}\multicolumn{14}{l}{\tabgrouplabel{WM-based Methods}} \\
World4Drive \cite{zheng2025world4drive} & \tabvenue{ICCV'25} & 3$\times$C & \tabno & \tabnum{97.8} & \tabnum{96.3} & \tabnum{99.4} & \tabsecond{99.8} & \tabnum{88.3} & \tabnum{97.1} & \tabbest{97.7} & \tabthird{98.0} & \tabnum{53.9} & \tabnum{84.8} \\
Epona \cite{zhang2025epona} & \tabvenue{ICCV'25} & 1$\times$C & \tabyes & \tabnum{97.1} & \tabnum{95.7} & \tabnum{99.3} & \tabthird{99.7} & \tabbest{88.6} & \tabnum{96.3} & \tabnum{97.0} & \tabthird{98.0} & \tabnum{67.8} & \tabnum{85.1} \\
Latent-WAM \cite{wang2026latent} & \tabvenue{arXiv'26} & 3$\times$C & \tabyes & \tabnum{98.1} & \tabthird{97.3} & \tabsecond{99.6} & \tabsecond{99.8} & \tabnum{87.7} & \tabnum{97.3} & \tabsecond{97.6} & \tabsecond{98.1} & \tabthird{87.3} & \tabsecond{89.3} \\
\midrule
\rowcolor{tabours}\textbf{CodeWAM} {\scriptsize(Ours)} & & 4$\times$C & \tabyes & \tabbest{98.9} & \tabsecond{99.0} & \tabbest{99.8} & \tabbest{99.9} & \tabsecond{88.5} & \tabbest{98.5} & \tabnum{97.2} & \tabthird{98.0} & \tabbest{87.9} & \tabbest{92.1} \\
\bottomrule
\end{tabularx}
\end{table*}

\begin{table*}[!tbp]
\centering
\caption{Two-stage planning on NAVSIM. S1 and S2 report stage-specific components; EPDMS is the overall score ($\uparrow$, \%). MV/SV: multi-/single-view cameras. Colors and superscripts indicate the top three values within each stage, and separately for EPDMS.}
\label{tab:navsim-two-stage}
\footnotesize
\setlength{\tabcolsep}{2pt}
\tabstyle
\begin{tabularx}{\textwidth}{l c c *{9}{>{\centering\arraybackslash}X} >{\bfseries}c}
\toprule
\rowcolor{tabheadbg}\textbf{Method} & \textbf{Sensors} & \textbf{Stage} & \tabhead{NC} & \tabhead{DAC} & \tabhead{DDC} & \tabhead{TLC} & \tabhead{EP} & \tabhead{TTC} & \tabhead{LK} & \tabhead{HC} & \tabhead{EC} & \tabhead{EPDMS} \\
\midrule
WorldDrive \cite{gui2026bridging} & SV & \tabstages{S1}{S2} & \tabstages{\tabbest{97.3}}{\tabbest{91.4}} & \tabstages{\tabsecond{89.1}}{\tabsecond{82.0}} & \tabstages{\tabnum{97.6}}{\tabsecond{91.0}} & \tabstages{\tabsecond{99.7}}{\tabsecond{98.5}} & \tabstages{\tabnum{60.5}}{\tabnum{53.1}} & \tabstages{\tabbest{96.8}}{\tabbest{90.6}} & \tabstages{\tabnum{87.7}}{\tabsecond{52.3}} & \tabstages{\tabnum{93.1}}{\tabnum{93.3}} & \tabstages{\tabnum{60.0}}{\tabthird{62.8}} & \tabsecond{34.9} \\[2pt]
\midrule
DiffusionDrive \cite{liao2025diffusiondrive} & MV & \tabstages{S1}{S2} & \tabstages{\tabsecond{96.8}}{\tabthird{80.1}} & \tabstages{\tabthird{86.0}}{\tabthird{72.8}} & \tabstages{\tabthird{98.8}}{\tabthird{84.4}} & \tabstages{\tabnum{99.3}}{\tabthird{98.4}} & \tabstages{\tabsecond{84.0}}{\tabbest{85.9}} & \tabstages{\tabthird{95.8}}{\tabthird{76.6}} & \tabstages{\tabsecond{96.7}}{\tabthird{46.4}} & \tabstages{\tabbest{97.6}}{\tabbest{96.3}} & \tabstages{\tabbest{79.6}}{\tabsecond{72.8}} & \tabthird{27.5} \\[2pt]
\midrule
LTF \cite{chitta2022transfuser} & MV & \tabstages{S1}{S2} & \tabstages{\tabthird{96.2}}{\tabnum{77.7}} & \tabstages{\tabnum{79.5}}{\tabnum{70.2}} & \tabstages{\tabsecond{99.1}}{\tabnum{84.2}} & \tabstages{\tabthird{99.5}}{\tabnum{98.0}} & \tabstages{\tabbest{84.1}}{\tabsecond{85.1}} & \tabstages{\tabnum{95.1}}{\tabnum{75.6}} & \tabstages{\tabthird{94.2}}{\tabnum{45.4}} & \tabstages{\tabsecond{97.5}}{\tabsecond{95.7}} & \tabstages{\tabsecond{79.1}}{\tabbest{75.9}} & \tabnum{23.1} \\[2pt]
\midrule
\rowcolor{tabours}\textbf{CodeWAM} {\scriptsize(Ours)} & MV & \tabstages{S1}{S2} & \tabstages{\tabnum{95.6}}{\tabsecond{89.9}} & \tabstages{\tabbest{96.4}}{\tabbest{88.1}} & \tabstages{\tabbest{100.0}}{\tabbest{94.4}} & \tabstages{\tabbest{100.0}}{\tabbest{98.9}} & \tabstages{\tabthird{80.4}}{\tabthird{80.7}} & \tabstages{\tabsecond{96.1}}{\tabsecond{86.4}} & \tabstages{\tabbest{97.1}}{\tabbest{60.8}} & \tabstages{\tabthird{96.9}}{\tabthird{94.5}} & \tabstages{\tabthird{75.2}}{\tabnum{58.6}} & \tabbest{50.6} \\[2pt]
\bottomrule
\end{tabularx}
\end{table*}

\subsection{Training the Scene Model}
\label{sec:method-training}

To cover states encountered during planning, we train on logged states, pose perturbations, planner-induced states, and rendered scenes.
BEV features at perturbed poses are resampled from the same images; planner-induced states are expressed in the matching log frame $H$ seconds later.
At each state, the current policy generates candidates as at deployment, and an offline simulator scorer provides their consequence labels.
We retain complete candidate sets for cross-candidate attention and perform two rounds of candidate regeneration and data aggregation after initialization on logged and perturbed states.
At deployment, the scene model predicts consequences without the scorer.

The policy and scene model share the visual encoder and are jointly trained with plan imitation, consequence prediction, and auxiliary BEV and map field losses.
The program $\mathcal P$ and consequence map $\mathcal R$ require no training.

\section{Experiments}
\label{sec:experiments}

\subsection{Experimental Setup}
\label{sec:setup}

\paragraph{Benchmarks and metrics.}
We evaluate CodeWAM on NAVSIM v1 navtest and NAVSIM v2 navtest, reporting PDMS and corrected EPDMS, respectively, with their component scores.
We also report two-stage EPDMS on the more challenging NAVHARD split.
For zero-shot transfer, we apply the NAVSIM-trained model to nuScenes.
\paragraph{Evaluation protocol.}
CodeWAM uses four camera views and takes ego status as input.
All methods in the comparison tables are listed with their publication venue, sensor configuration, and whether ego status is used.
Ranking colors are assigned per metric across all listed methods, including CodeWAM; equal displayed values share the same rank, and the next distinct value receives the next rank.
All models are trained on 8 NVIDIA H20 GPUs.

\subsection{Results}
\label{sec:results}

Tables~\ref{tab:navsim-v1}--\ref{tab:nuscenes} compare CodeWAM with published methods, grouped into end-to-end, VLM-based, and world-model-based approaches where applicable.

\begin{table}[!tbp]
\caption{Open-loop planning on nuScenes validation; CodeWAM is zero-shot. L2 (m) is averaged up to each horizon ($\downarrow$). Colors and superscripts mark the top three distinct values; ties share a rank.}
\label{tab:nuscenes}
\centering
\footnotesize
\setlength{\tabcolsep}{1.5pt}
\tabstyle
\begin{tabularx}{\columnwidth}{l l<{\hspace{1pt}} l<{\hspace{1pt}} c *{3}{>{\centering\arraybackslash}X} >{\bfseries\centering\arraybackslash}X}
\toprule
\multicolumn{4}{c}{\textbf{}} & \multicolumn{4}{c}{\textbf{L2 (m) $\downarrow$}} \\
\cmidrule(lr){5-8}
\rowcolor{tabheadbg}\textbf{Method} & \textbf{Venue} & \textbf{Sensors} & \textbf{Ego} & \tabhead{1s} & \tabhead{2s} & \tabhead{3s} & \tabhead{Avg.} \\
\midrule
UniAD \cite{hu2023planning} & \tabvenue{CVPR} & 6$\times$C & \tabyes & \tabnum{0.48} & \tabnum{0.96} & \tabnum{1.65} & \tabnum{1.03} \\
VAD-Base \cite{jiang2023vad} & \tabvenue{ICCV} & 6$\times$C & \tabyes & \tabnum{0.41} & \tabnum{0.70} & \tabnum{1.05} & \tabnum{0.72} \\
GenAD \cite{zheng2024genad} & \tabvenue{ECCV} & 6$\times$C & \tabyes & \tabthird{0.36} & \tabnum{0.83} & \tabnum{1.55} & \tabnum{0.91} \\
VLP \cite{pan2024vlp} & \tabvenue{CVPR} & 6$\times$C & \tabyes & \tabbest{0.30} & \tabsecond{0.53} & \tabsecond{0.84} & \tabsecond{0.55} \\
ReAL-AD \cite{lu2025real} & \tabvenue{ICCV} & 6$\times$C & \tabyes & \tabbest{0.30} & \tabbest{0.48} & \tabbest{0.67} & \tabbest{0.48} \\
Drive-WM \cite{wang2024driving} & \tabvenue{CVPR} & 6$\times$C & \tabno & \tabnum{0.43} & \tabnum{0.77} & \tabnum{1.20} & \tabnum{0.80} \\
\midrule
\rowcolor{tabours}\textbf{CodeWAM} {\scriptsize(Ours)} & & 4$\times$C & \tabyes & \tabsecond{0.34} & \tabthird{0.62} & \tabthird{0.96} & \tabthird{0.64} \\
\bottomrule
\end{tabularx}
\end{table}

\begin{figure}[!tbp]
\centering
\includegraphics[width=\columnwidth]{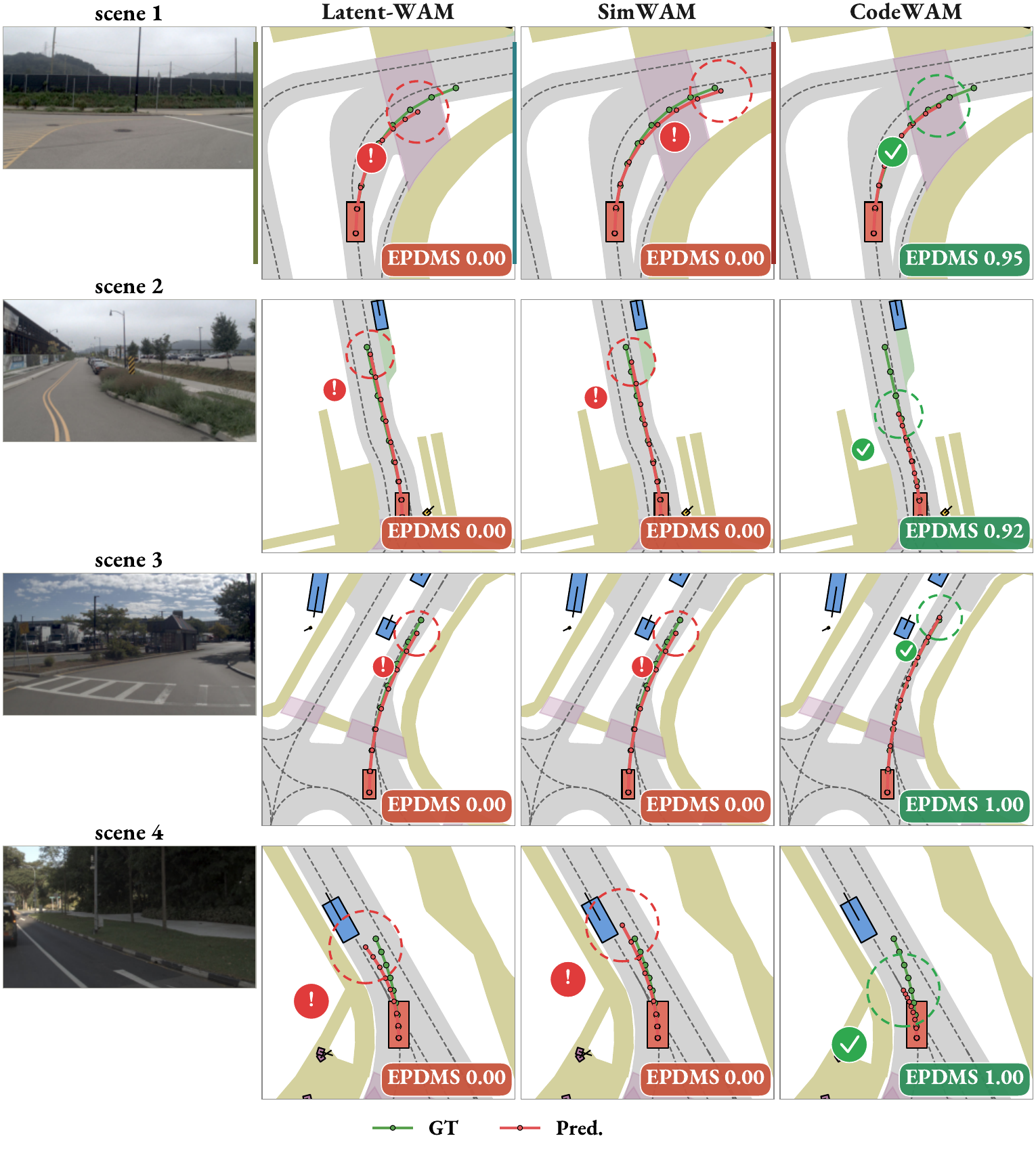}
\caption{Qualitative planning on NAVSIM v2 navtest. Each row shows a front-camera observation and bird's-eye plans from Latent-WAM, SimWAM, and CodeWAM. Green denotes the recorded human trajectory and red the predicted plan; badges report frame EPDMS on $[0,1]$. CodeWAM obtains nonzero scores in all four selected cases, where both baselines score zero.}
\label{fig:tri_main}
\end{figure}

\begin{figure}[!tbp]
\centering
\includegraphics[width=\columnwidth]{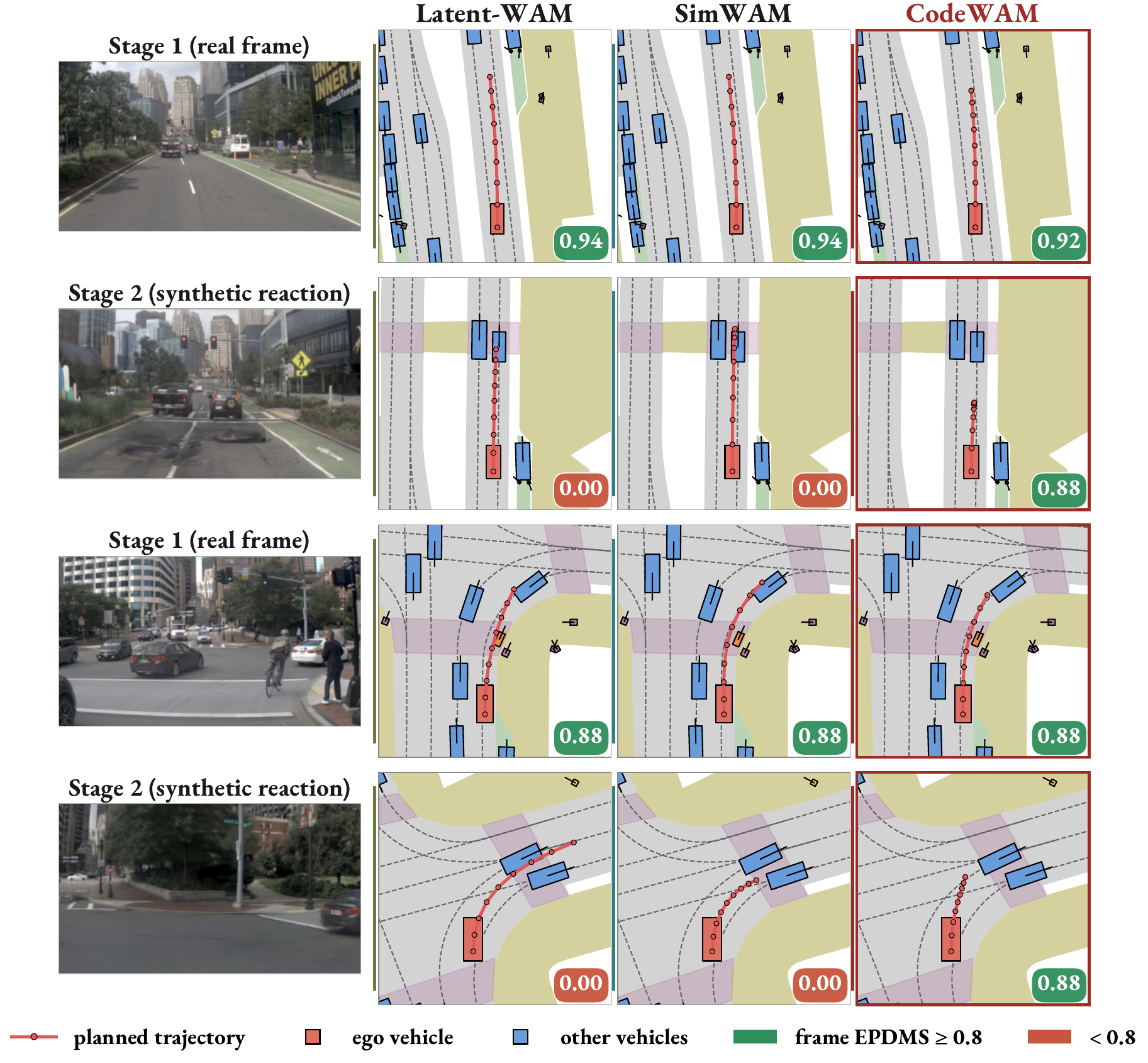}
\caption{Two-stage decisions on NAVHARD. Each pair of rows shows a real stage-1 observation followed by a synthetic stage-2 observation, with plans from Latent-WAM, SimWAM, and CodeWAM. Red denotes planned motion. Badges report frame EPDMS, with green indicating $\geq 0.8$ and orange lower values. CodeWAM retains nonzero stage-2 scores in both examples; these badges are not combined two-stage scores.}
\label{fig:nh_stage_main}
\end{figure}

\paragraph{NAVSIM planning.}
CodeWAM achieves 93.6 PDMS on NAVSIM v1 navtest and 92.1 EPDMS on v2, exceeding the strongest compared baselines by 1.9 and 2.8 percentage points, respectively (Tables~\ref{tab:navsim-v1}--\ref{tab:navsim-v2}).
On v1, it combines the highest progress score (90.2) with 99.6 NC; on v2, it leads in NC, DDC, TLC, TTC, and EC, demonstrating strong safety, compliance, and cross-plan comfort.
Under two-stage evaluation, CodeWAM reaches 50.6 EPDMS, compared with 34.9 for WorldDrive (Table~\ref{tab:navsim-two-stage}), and ranks first in DAC, DDC, TLC, and LK at both stages.
Its stage-2 DAC and LK reach 88.1 and 60.8, exceeding the next-best values by 6.1 and 8.5 points, respectively.
Figures~\ref{fig:tri_main} and~\ref{fig:nh_stage_main} complement these results with representative trajectory comparisons against Latent-WAM and SimWAM on navtest and NAVHARD, including the outcomes of planning stages.

\begin{table}[!tbp]
\centering
\caption{Candidate-count study with frozen DrivoR proposals. EPDMS follows corrected NAVHARD two-stage evaluation; pairwise EC is diagnostic. A dash denotes missing data.}
\label{tab:candidate-count}
\footnotesize
\setlength{\tabcolsep}{3pt}
\tabstyle
\begin{tabularx}{\columnwidth}{c *{4}{>{\centering\arraybackslash}X}}
\toprule
 & \multicolumn{2}{c}{\tabhead{EPDMS $\uparrow$}} & \multicolumn{2}{c}{\tabhead{Pairwise EC (\%) $\uparrow$}} \\
\cmidrule(lr){2-3}\cmidrule(lr){4-5}
\rowcolor{tabheadbg}\tabhead{$k$} & \tabhead{Learned} & \tabhead{+ Exec.} & \tabhead{Learned} & \tabhead{+ Exec.} \\
\midrule
1  & \tabnum{39.99} & \tabnum{39.99} & \tabnum{45.2} & \tabnum{45.2} \\
2  & \tabnum{44.46} & \tabnum{44.84} & -- & -- \\
4  & \tabnum{44.56} & \tabnum{45.04} & \tabnum{43.5} & \tabnum{47.9} \\
8  & \tabnum{50.96} & \tabnum{52.48} & \tabnum{42.9} & \tabnum{50.3} \\
16 & \tabnum{54.19} & \tabnum{56.34} & \tabnum{42.8} & \tabnum{54.1} \\
32 & \tabnum{54.43} & \tabnum{55.67} & \tabnum{41.7} & \tabnum{56.3} \\
\rowcolor{tabours}64 & \tabnum{54.56} & \tabnum{56.70} & \tabnum{40.2} & \tabnum{57.8} \\
\bottomrule
\end{tabularx}
\end{table}

\begin{figure}[!tbp]
\centering
\includegraphics[width=\columnwidth]{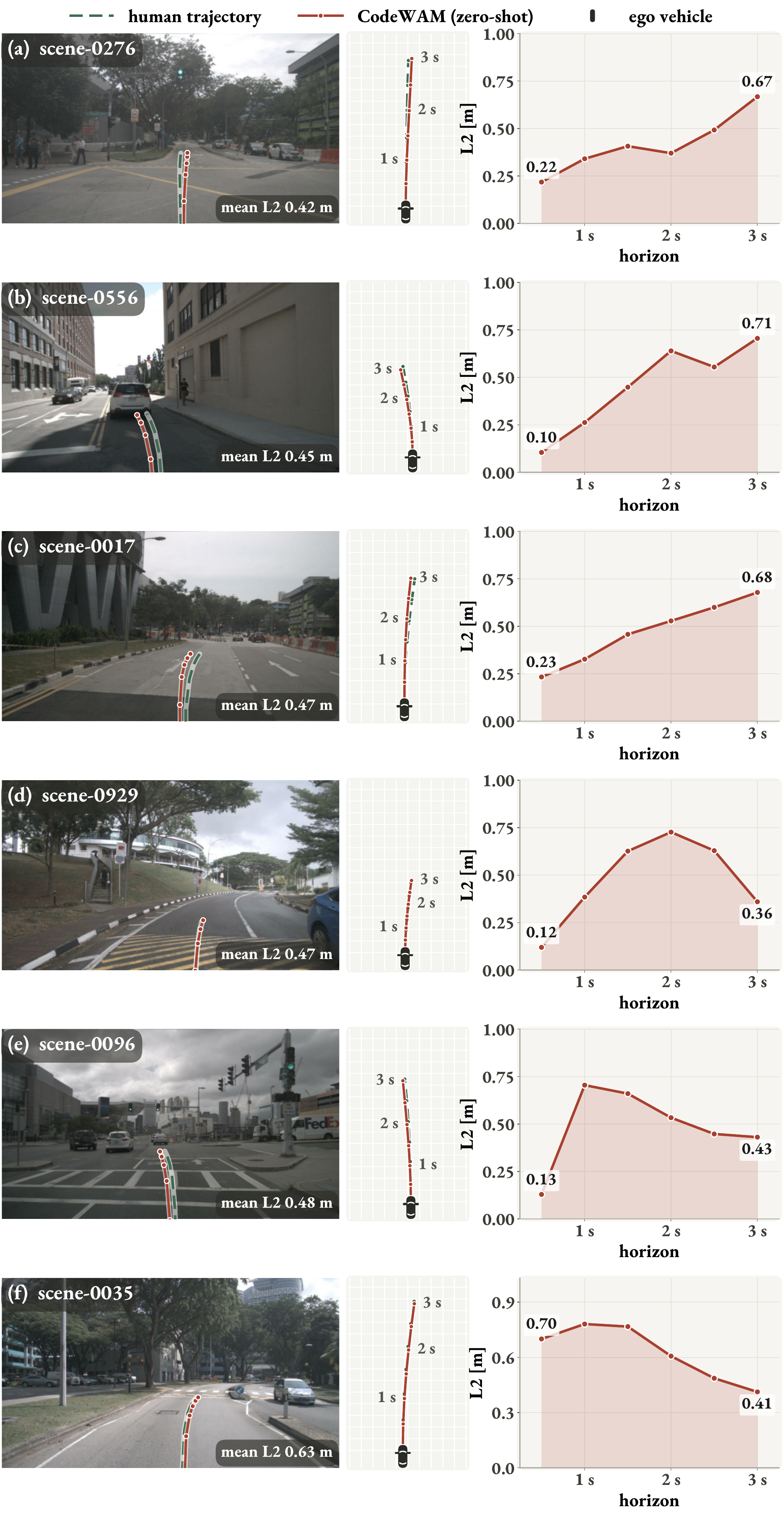}
\caption{Zero-shot transfer to nuScenes without fine-tuning. Each row shows camera and bird's-eye views of CodeWAM's prediction (red) and recorded human trajectory (dashed green), followed by pointwise L2 error in meters at 0.5-second intervals up to 3\,s. Image badges average these six errors; they do not report the average of the three horizon-averaged metrics in Table~\ref{tab:nuscenes}.}
\label{fig:nusc_main}
\end{figure}

\paragraph{Zero-shot transfer to nuScenes.}
Without any fine-tuning, the NAVSIM-trained CodeWAM transfers to real nuScenes data with competitive L2 (Table~\ref{tab:nuscenes}).
Figure~\ref{fig:nusc_main} compares the predicted trajectory with recorded human motion across diverse scenes and shows how pointwise displacement error changes over the prediction horizon.

\subsection{Qualitative Analysis}
\label{sec:qualitative}

\begin{figure}[!tbp]
\centering
\includegraphics[width=\columnwidth]{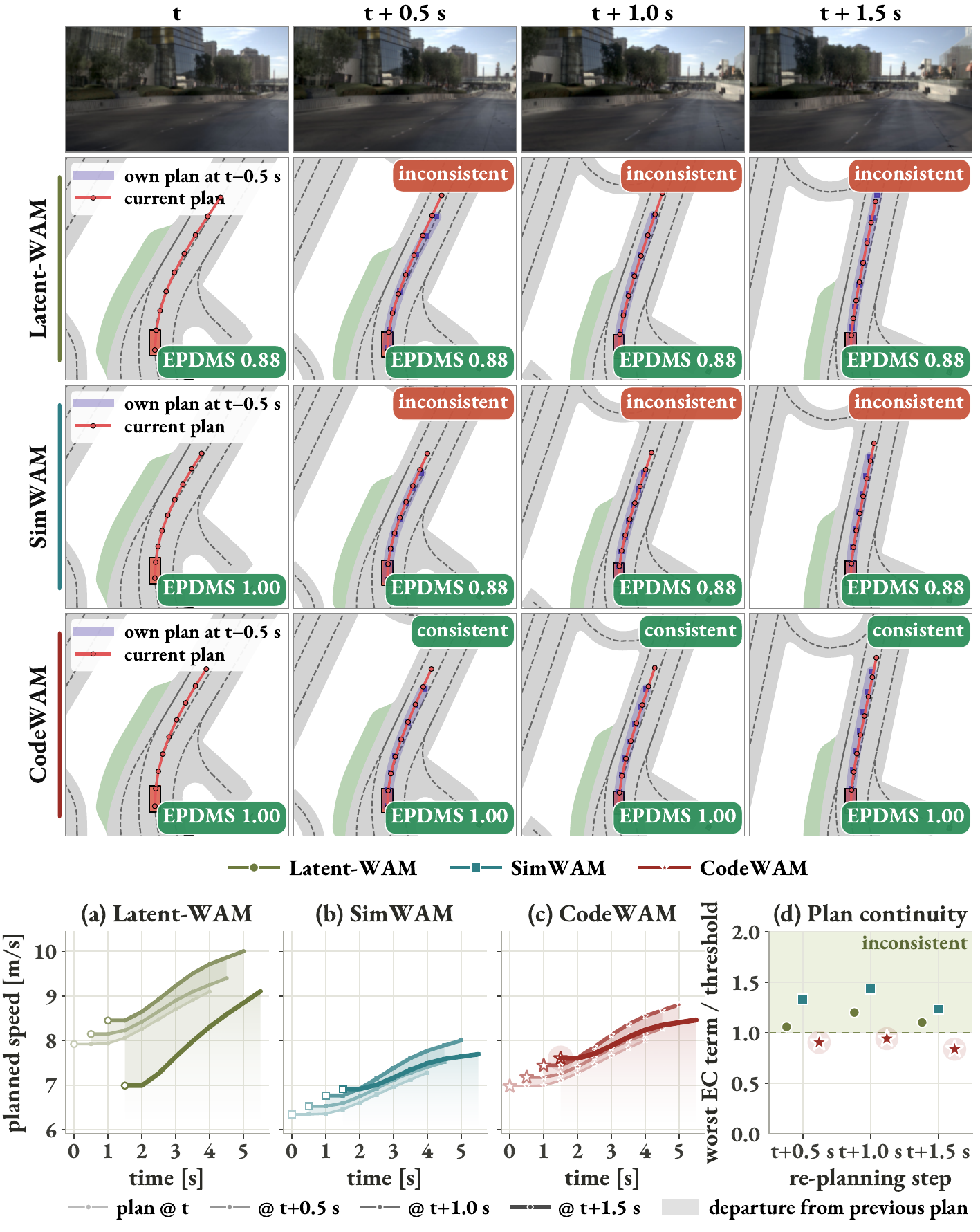}
\caption{Consecutive decisions at 0.5-second intervals. Rows compare Latent-WAM, SimWAM, and CodeWAM; red shows the current plan and purple each method's preceding plan. Badges report frame EPDMS. Below, planned speed profiles accompany the largest threshold-normalized EC difference, where values above one indicate inconsistency. CodeWAM passes all three continuity checks, while both baselines fail them.}
\label{fig:strip_f}
\end{figure}

Figure~\ref{fig:strip_f} follows consecutive planning decisions along a driving sequence, with replanning at 0.5-second intervals.
CodeWAM carries the preceding plan forward as the reference for execution evaluation. Its planned speed profiles remain compatible across all three updates, while both baselines exceed the displayed EC threshold. These plots describe consistency between predicted plans, rather than measured ego motion from a closed-loop rollout.

\begin{table}[!tbp]
\centering
\caption{Execution evaluation without retraining: corrected NAVHARD two-stage EPDMS ($\uparrow$). Gains are score-point differences relative to the original planner.}
\label{tab:planner-integration}
\footnotesize
\setlength{\tabcolsep}{3pt}
\tabstyle
\begin{tabularx}{\columnwidth}{l *{3}{>{\centering\arraybackslash}X}}
\toprule
\rowcolor{tabheadbg}\tabhead{Planner} & \tabhead{Original} & \tabhead{+ Exec.} & \tabhead{Gain} \\
\midrule
Hydra-MDP & \tabnum{41.16} & \tabnum{43.35} & \textbf{+2.19} \\
DrivoR & \tabnum{54.56} & \tabnum{56.70} & \textbf{+2.14} \\
\bottomrule
\end{tabularx}
\end{table}

\subsection{Analysis of Execution Evaluation}
\label{sec:execution-studies}

\paragraph{Candidate count and execution consistency.}
We vary the selectable subset of a fixed pool of 64 DrivoR proposals~\cite{kirby2026driving}, keeping the model frozen.
Learned scoring selects the highest-scoring candidate; + Execution adds the program-computed EC term with weight 6, without training.
EPDMS uses all 5{,}912 evaluation frames.
Pairwise EC measures two-frame consistency on a separate 2{,}956-frame diagnostic subset, with previous-plan references fixed to the learned-scoring selections; it is not the official stage-2 EC.
For $k\in\{1,4,8,16,32\}$, this diagnostic averages 20 random subset seeds; $k=64$ uses the full pool.
As $k$ increases from 1 to 64, learned scoring improves EPDMS but reduces pairwise EC from 45.2\% to 40.2\% (Table~\ref{tab:candidate-count}).
At $k=64$, execution evaluation raises pairwise EC to 57.8\% (+17.6 points) and EPDMS from 54.56 to 56.70 (+2.14).
The variants coincide at $k=1$, where re-selection is impossible.

\paragraph{Integration with existing planners.}
We add execution evaluation to candidate selection in Hydra-MDP and DrivoR without additional training (Table~\ref{tab:planner-integration}).
EPDMS improves from 41.16 to 43.35 for Hydra-MDP (+2.19) and from 54.56 to 56.70 for DrivoR (+2.14); the latter reuses the $k=64$ setting above.
These results support transfer across candidate-scoring planners beyond CodeWAM, while search-based and single-trajectory planners remain untested here.

\section{Conclusion}
We presented CodeWAM, a driving world model that uses executable code as part of its world representation. Code models ego dynamics under candidate plans, while a visual model predicts scene-dependent consequences. By evaluating each candidate against the prior plan and recorded motion history, CodeWAM links successive decisions through a shared execution context. Experiments show strong planning performance, improved execution consistency, and gains on existing planners without additional training. These results support executable code as a complement to learned scene models, combining explicit vehicle dynamics with scene understanding for planning.

{
    \small
    \bibliographystyle{ieeenat_fullname}
    \bibliography{main}
}

\clearpage
\setcounter{page}{1}
\renewcommand{\thepage}{S\arabic{page}}
\maketitlesupplementary

\appendix
\setcounter{equation}{0}
\renewcommand{\theequation}{A\arabic{equation}}
\renewcommand{\theHequation}{supp.\arabic{equation}}
\section{Benchmarks and Evaluation Metrics}
\label{sec:supp-metrics}

We describe the evaluation settings and aggregation rules for five benchmarks. Component scores lie in $[0,1]$, with larger values indicating better performance. Driving scores are scaled by 100 for percentage reporting; nuScenes L2 errors are in meters and are lower-is-better. Composite scores are calculated before dataset aggregation.

\subsection{NAVSIM v1: PDM Score}
\label{sec:supp-pdms}

NAVSIM v1 evaluates a predicted plan using a four-second non-reactive simulation: an LQR tracker and a kinematic bicycle model determine ego motion, while surrounding actors follow their recorded trajectories. The Predictive Driver Model Score (PDMS) combines safety constraints with progress, collision margin, and comfort:
\begin{equation}
\mathrm{PDMS}
= \mathrm{NC} \cdot \mathrm{DAC} \cdot
\frac{5\,\mathrm{EP} + 5\,\mathrm{TTC} + 2\,\mathrm{C}}{12}.
\label{eq:supp-pdms}
\end{equation}

\paragraph{Safety constraints.}
No at-fault Collisions (NC) assigns 1 without a responsible collision, 0 for an at-fault collision with a traffic participant, and 0.5 for one with a non-agent object. Drivable Area Compliance (DAC) is 1 only if the footprint stays within the drivable area. A zero in either factor makes PDMS zero.

\paragraph{Weighted driving quality.}
Ego Progress (EP) measures route-centerline progress, normalized using the PDM reference planner and clipped to $[0,1]$. Time to Collision within bound (TTC) returns 0 for a safety-margin violation under short-horizon collision extrapolation. Comfort (C) tests acceleration, jerk, yaw rate, and yaw acceleration against benchmark limits. Reported PDMS averages per-scene scores, not separately averaged components.

\subsection{NAVSIM v2: Extended PDM Score}
\label{sec:supp-epdms}

NAVSIM v2 extends PDMS with additional rule-compliance and temporal-comfort criteria. Our navtest results use the single-stage non-reactive setting. The four multiplicative metrics are
$\mathcal M=\{\mathrm{NC},\mathrm{DAC},\mathrm{DDC},\mathrm{TLC}\}$,
and the weighted metrics are
$\mathcal W=\{\mathrm{EP},\mathrm{TTC},\mathrm{LK},\mathrm{HC},\mathrm{EC}\}$.

\paragraph{Additional components.}
Driving Direction Compliance (DDC) penalizes travel in oncoming lanes outside intersections. With the default one-second window, the score is 1, 0.5, or 0 for oncoming travel below 2\,m, from 2\,m to below 6\,m, or at least 6\,m, respectively. Traffic Light Compliance (TLC) checks that the rollout does not cross the red-light regions represented by the evaluator. Lane Keeping (LK) penalizes sustained lateral departure from the route centerline; the default limits are 0.5\,m for 2\,s, with intersection samples excluded. History Comfort (HC) evaluates comfort on the rollout concatenated with recorded past motion, thereby also testing the transition at the planning boundary.

Extended Comfort (EC) tests consistency between simulated plans from adjacent observations. After temporal alignment, let $q_{t,j}$ and $q_{t-1,j}^{\mathrm{aligned}}$ denote corresponding samples over their overlap $\mathcal J$. For each motion feature $q$,
\begin{equation}
\begin{aligned}
d_q&=\sqrt{\frac{1}{|\mathcal J|}\sum_{j\in\mathcal J}
\bigl(q_{t,j}-q_{t-1,j}^{\mathrm{aligned}}\bigr)^2},\\
\mathrm{EC}&=\prod_{q\in\mathcal Q}\mathbf{1}\!\left[d_q\le b_q\right].
\end{aligned}
\label{eq:supp-ec}
\end{equation}
The features $\mathcal Q$ are acceleration magnitude, jerk magnitude, yaw rate, and yaw acceleration, with RMS limits of 0.7\,m/s$^2$, 0.5\,m/s$^3$, 0.1\,rad/s, and 0.1\,rad/s$^2$. Table~\ref{tab:candidate-count}'s fixed-reference pairwise EC diagnostic is separate from official stage-specific EC.

\paragraph{Human-reference filtering.}
The corrected evaluator suppresses eligible penalties when the simulated human reference receives zero for the same component. Where this test applies, write
\begin{equation}
\widetilde m=
\begin{cases}
1, & \text{if }m^{\mathrm{human}}=0,\\
m, & \text{otherwise}.
\end{cases}
\label{eq:supp-human-filter}
\end{equation}
Filtering applies to original recorded scenes, not synthetic continuations or the subsequently computed EC. The corrected evaluator updates both multiplicative and weighted terms from the filtered components.

\paragraph{Score calculation.}
When EC is available, the Extended PDM Score (EPDMS) is
\begin{equation}
\begin{aligned}
P&=\prod_{m\in\mathcal M}\widetilde m,\\
Q&=5\,\widetilde{\mathrm{EP}}+5\,\widetilde{\mathrm{TTC}}
  +2\,\widetilde{\mathrm{LK}}+2\,\widetilde{\mathrm{HC}},\\
\mathrm{EPDMS}&=P\cdot\frac{Q+2\,\mathrm{EC}}{16}.
\end{aligned}
\label{eq:supp-epdms}
\end{equation}
In single-stage evaluation, some observations lack a valid adjacent frame. The official evaluator removes EC and its weight for those observations, giving $PQ/14$, rather than assigning a failed EC test. Single-stage navtest aggregates the resulting per-scene scores by their arithmetic mean.

\subsection{NAVHARD: Two-Stage Pseudo-Simulation}
\label{sec:supp-navhard}

NAVHARD uses two-stage pseudo-simulation with reactive traffic. Stage 1 uses recorded observations; stage 2 uses pre-rendered observations at alternative future ego states. Each evaluates a four-second plan with EPDMS. Continuations are weighted by their proximity to the first-stage endpoint, testing recovery from states the planner is likely to encounter.

Let $p_i$ be the first-stage endpoint for evaluation branch $i$, and let $z_{ij}$ be the start position of its $j$-th second-stage continuation. The official proximity weights are
\begin{equation}
\begin{aligned}
\kappa_{ij}&=\exp\!\left(-\frac{\lVert p_i-z_{ij}\rVert_2^2}{2\sigma^2}\right),\\
w_{ij}&=\frac{\kappa_{ij}}{\sum_{\ell}\kappa_{i\ell}}.
\end{aligned}
\label{eq:supp-navhard-weights}
\end{equation}
The released aggregator uses $\sigma^2=0.1\,\mathrm{m}^2$ and falls back to uniform weights when the kernel sum is numerically zero. With $E_i^{(1)}$ and $E_{ij}^{(2)}$ denoting normalized stage scores, the branch score is
\begin{equation}
S_i=E_i^{(1)}\cdot\sum_j w_{ij}\,E_{ij}^{(2)}.
\label{eq:supp-navhard-branch}
\end{equation}
For each of $G$ official paired-frame groups, the evaluator computes the branch score for both the current and preceding frame, averages these two scores, and then averages over groups:
\begin{equation}
\mathrm{EPDMS}_{\mathrm{NAVHARD}}
=\frac{100}{G}\,\sum_{g=1}^{G}\frac{S_{g,\mathrm{now}}+S_{g,\mathrm{prev}}}{2}.
\label{eq:supp-navhard}
\end{equation}
This combined score differs from averaging all frames or multiplying dataset-level stage means. S1 and S2 components summarize their respective stages.

\subsection{nuScenes: Open-Loop Planning Error}
\label{sec:supp-nuscenes}

nuScenes open-loop evaluation compares predicted ego positions with recorded motion. CodeWAM transfers from NAVSIM without nuScenes fine-tuning; predicted plans do not alter subsequent observations.

Let $\widehat p_{n,k},p_{n,k}\in\mathbb R^2$ be predicted and ground-truth positions in a common coordinate frame, for sample $n$ and future time $t_k$. Following the horizon-averaged convention stated in Table~\ref{tab:nuscenes}, define
\begin{equation}
\begin{aligned}
e_k&=\frac{\sum_n v_{n,k}\,\lVert\widehat p_{n,k}-p_{n,k}\rVert_2}
{\sum_n v_{n,k}},\\[2pt]
\mathrm{L2}(h)&=\frac{1}{K_h}\,\sum_{k:t_k\le h}e_k,\\[2pt]
\mathrm{Avg.\ L2}&=\frac{\mathrm{L2}(1\,\mathrm{s})+\mathrm{L2}(2\,\mathrm{s})+\mathrm{L2}(3\,\mathrm{s})}{3},
\end{aligned}
\label{eq:supp-nuscenes-l2}
\end{equation}
Here $v_{n,k}$ marks valid ground truth and $K_h$ counts samples up to $h$: 2, 4, and 6 at 2\,Hz. Horizon averaging differs from endpoint-only error. L2 measures agreement with recorded motion, not closed-loop driving quality.

\subsection{HUGSIM: Route Completion and HD-Score}
\label{sec:supp-hugsim}

HUGSIM renders driving scenes with 3D Gaussian Splatting and updates observations after ego actions in a closed loop. We describe Route Completion (RC) and HD-Score as defined by its native scorer.

\paragraph{Route completion.}
RC measures progress along the prescribed route before termination. In the released environment, let $j_t$ be the zero-based index of the reference camera pose nearest to the ego position at step $t$, and let $L$ be the number of reference poses. The episode-level completion ratio is
\begin{equation}
R_c=\min\!\left\{1,\;\max_t\frac{j_t+1}{0.9\,L}\right\}.
\label{eq:supp-hugsim-rc}
\end{equation}
This is the native reference-index convention, including its 90\% completion normalization; it is not an arc-length ratio. RC is reported as $100R_c$.

\paragraph{Driving quality.}
For each evaluated planning step $t$, the native scorer checks the predicted trajectory against scene geometry and surrounding actors. NC tests collisions with foreground actors and background geometry. DAC checks coverage of the vehicle footprint by drivable-ground points. TTC tests collisions after short-horizon motion extrapolation, while comfort (COM) checks bounds on longitudinal and lateral acceleration, longitudinal jerk, yaw rate, and yaw acceleration. These terms have benchmark-specific implementations even where their names match NAVSIM components. The step score and episode HD-Score are
\begin{equation}
\begin{aligned}
q_t&=\mathrm{NC}_t\cdot\mathrm{DAC}_t\cdot
\frac{5\,\mathrm{TTC}_t+2\,\mathrm{COM}_t}{7},\\[2pt]
\text{HD-Score}&=100\,R_c\cdot\frac{1}{T}\sum_{t=1}^{T}q_t,
\end{aligned}
\label{eq:supp-hugsim}
\end{equation}
Here $T$ counts evaluated plans. Progress enters through episode-level RC. The native quality term evaluates issued plans, while observations evolve in a closed loop. Overall RC and HD-Score average episode values, weighting difficulty subsets by their episode counts.
\clearpage

\section{Additional Qualitative Analysis}
\label{sec:supp-qualitative}

We examine how scene prediction, execution evaluation, and candidate refinement affect individual decisions, and how those decisions connect over time. The following visualizations complement the quantitative results with selected cases. EPDMS badges inside a panel are normalized scores for that frame; they are distinct from the dataset-level percentages below the variant comparisons and from the combined two-stage NAVHARD score.

\subsection{Roles of the Planning Components}
\label{sec:supp-variants}

\begin{figure*}[!tp]
\centering
\includegraphics[width=\textwidth,height=0.84\textheight,keepaspectratio]{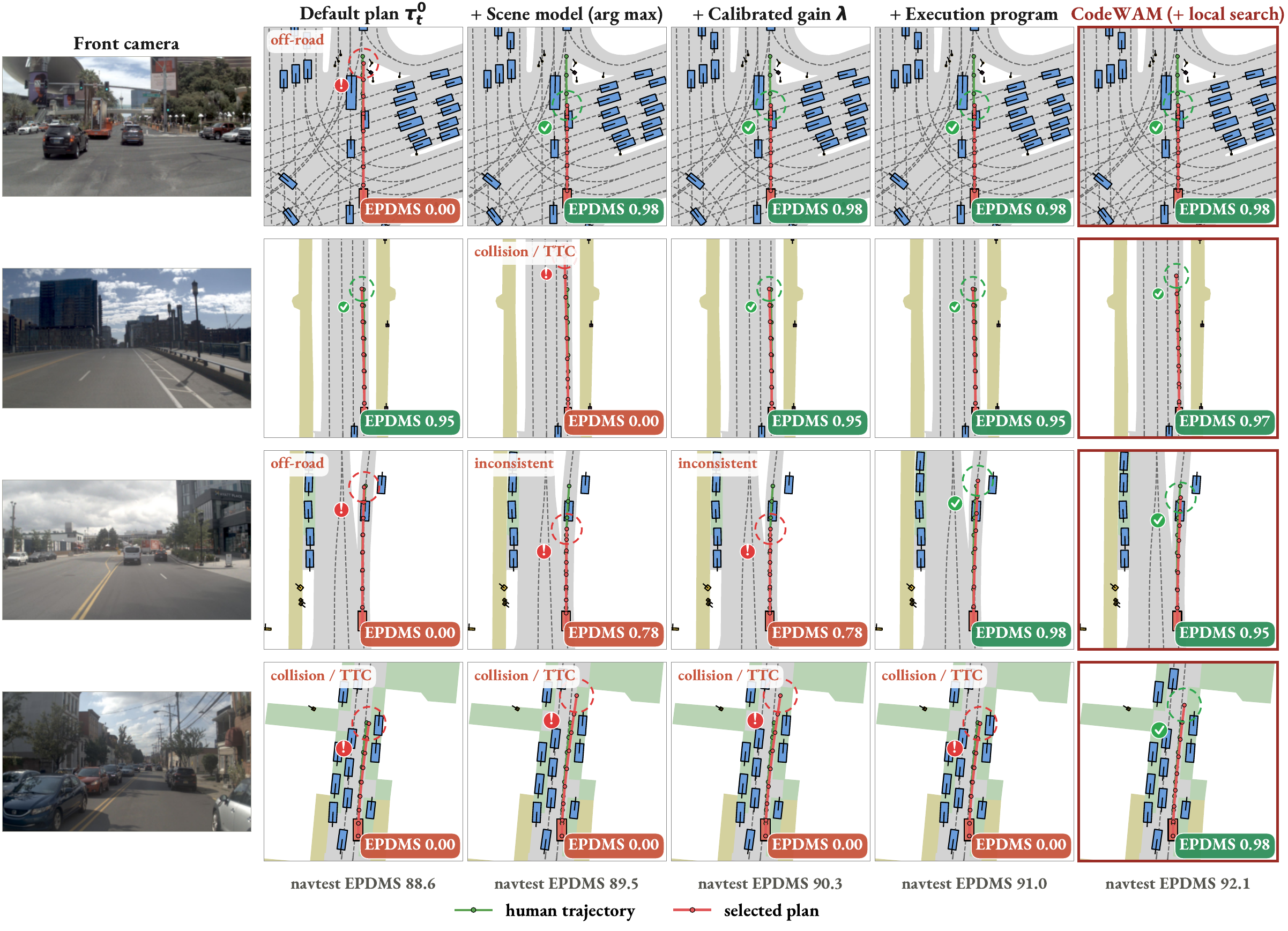}
\caption{Effects of successive planning components on NAVSIM v2 navtest. After the front-camera view, columns show the default policy plan, scene-score maximization, calibrated gain, the execution program, and local search, with components added cumulatively. From top to bottom, the examples show correction of an off-road proposal, rejection of an overrated alternative, resolution of execution inconsistency, and recovery through refinement. Green denotes recorded human motion and red the selected plan. Badges report frame EPDMS on $[0,1]$; footer values report dataset-level navtest EPDMS in percent.}
\label{fig:supp-versions-main}
\end{figure*}

Figure~\ref{fig:supp-versions-main} separates the contributions of scene scoring, calibrated selection, the execution program, and local search. In the first row, scene scoring replaces an off-road default plan with a valid alternative. The second row shows the opposite situation: unconstrained score maximization replaces a valid default with a candidate penalized for collision or TTC, while the calibrated gain preserves the default. The third row exposes an execution mismatch that scene scoring and calibration leave unresolved; adding the execution program raises the frame score from 0.785 to 0.978. The full model scores 0.951 on this case, illustrating that successive components need not improve every individual frame. In the last row, local search finds a plan with score 0.975 after the preceding variants receive zero.

\begin{figure*}[!tp]
\centering
\includegraphics[width=\textwidth,height=0.84\textheight,keepaspectratio]{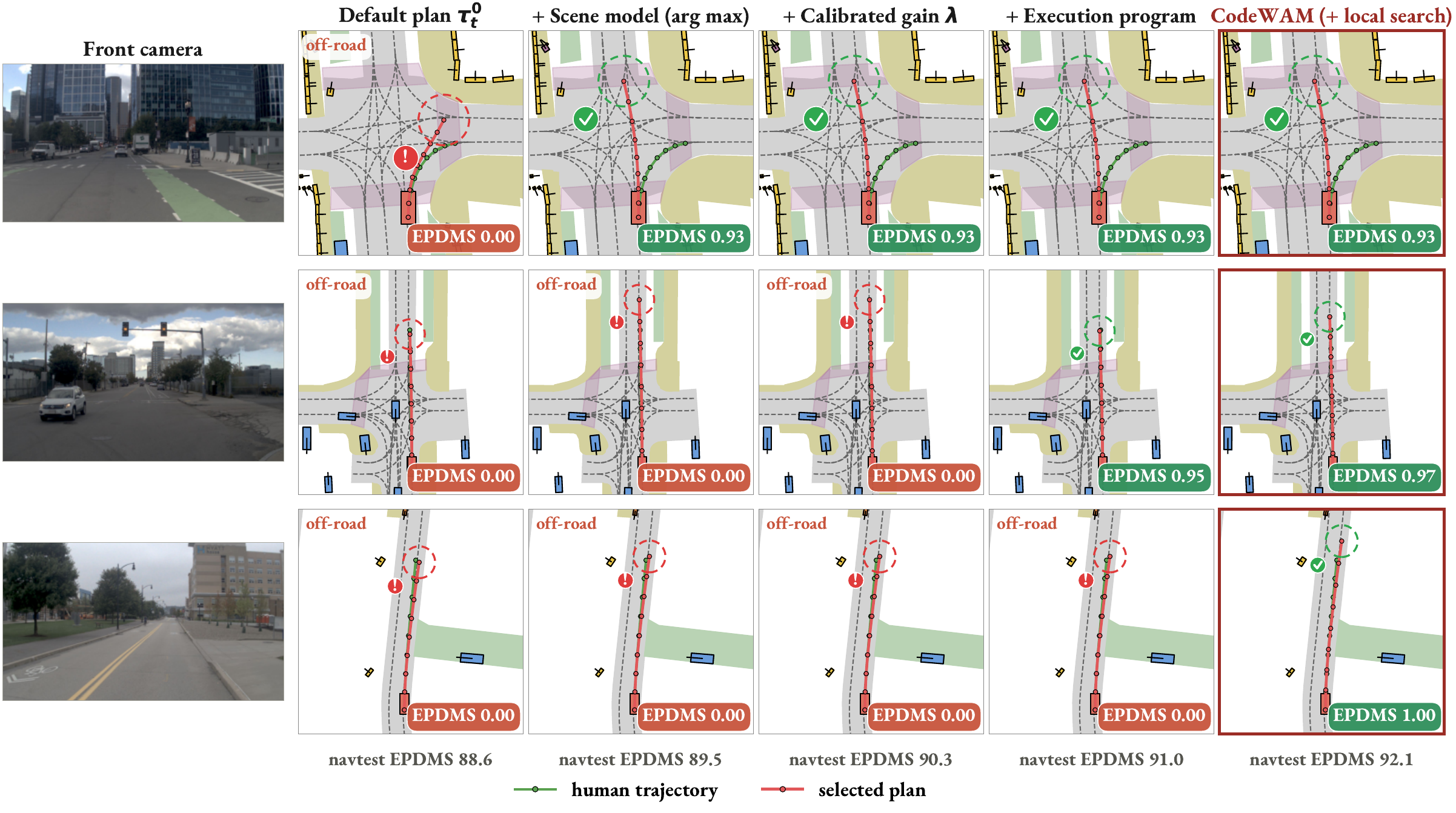}
\caption{Examples resolved at different steps of the planning procedure. Scene scoring first produces a nonzero frame score in the top row, the execution program in the middle row, and local search in the bottom row. Refinement further improves the middle-row score from 0.95 to 0.97. Column order, trajectory colors, and score conventions follow Figure~\ref{fig:supp-versions-main}; footer values summarize navtest performance rather than these three examples.}
\label{fig:supp-versions-progressive}
\end{figure*}

\begin{figure*}[!tp]
\centering
\includegraphics[width=\textwidth,height=0.84\textheight,keepaspectratio]{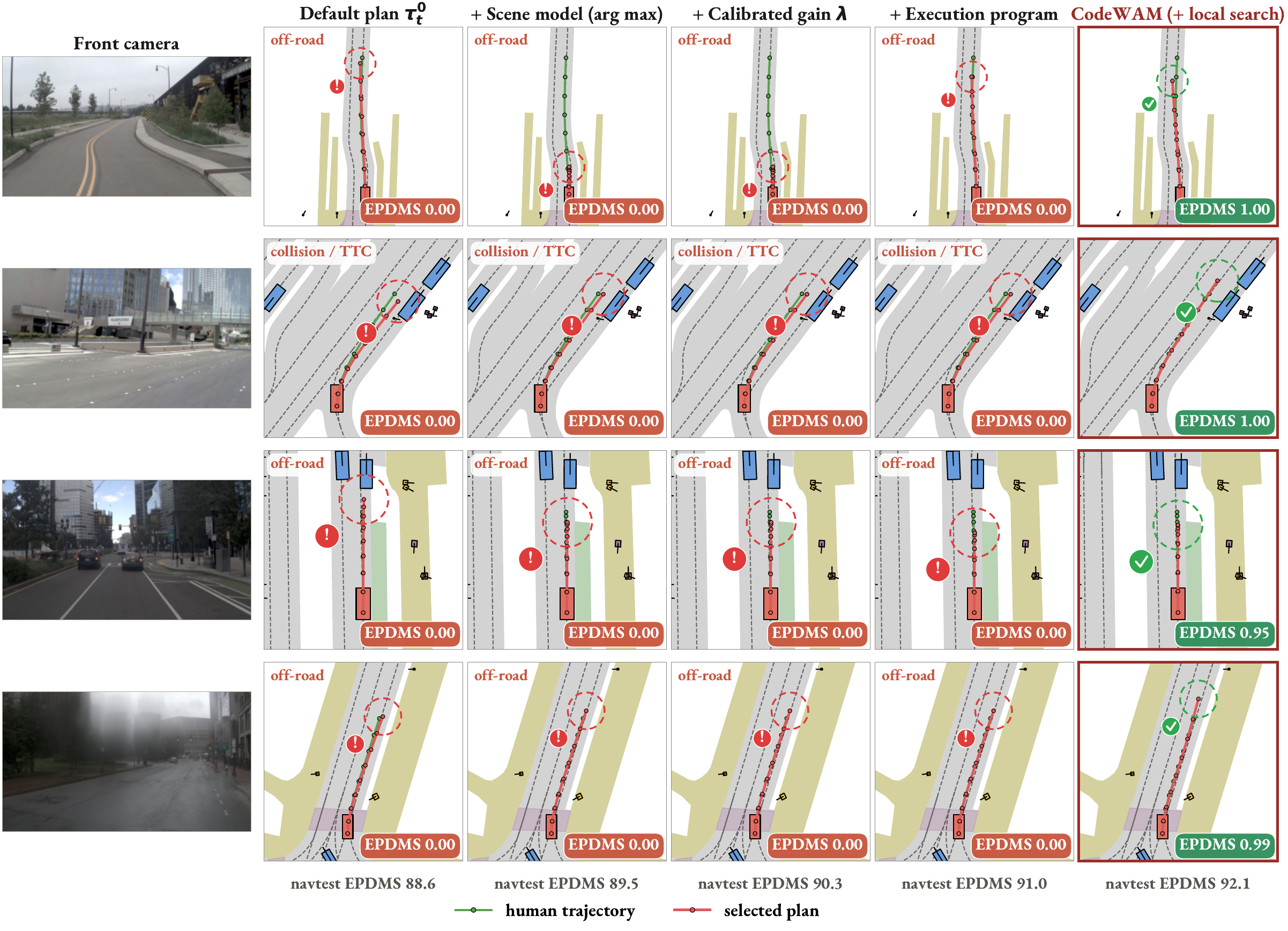}
\caption{Cases benefiting from local candidate refinement. The four variants preceding local search receive zero frame EPDMS, with off-road annotations in rows 1, 3, and 4 and collision/TTC annotations in row 2. Full CodeWAM obtains 1.00, 1.00, 0.95, and 0.99, respectively. These selected cases illustrate how refinement can recover from failures that remain after candidate re-selection. Column order, trajectory colors, and frame-versus-dataset score conventions follow Figure~\ref{fig:supp-versions-main}.}
\label{fig:supp-versions-hard}
\end{figure*}

Figure~\ref{fig:supp-versions-progressive} gives further examples in which scene evaluation, execution evaluation, or local refinement first produces a nonzero score. Figure~\ref{fig:supp-versions-hard} focuses on cases where re-selection alone remains insufficient: all four preceding variants receive zero, whereas the full model obtains scores between 0.946 and 1.000. Together, these cases illustrate why evaluating candidates and expanding the candidate set serve complementary roles. The execution context informs which motions are compatible with the current decision, while refinement can provide an option absent from the original selection.

\subsection{Candidate Selection under the Execution Context}
\label{sec:supp-candidates}

\begin{figure*}[!tp]
\centering
\includegraphics[width=\textwidth,height=0.84\textheight,keepaspectratio]{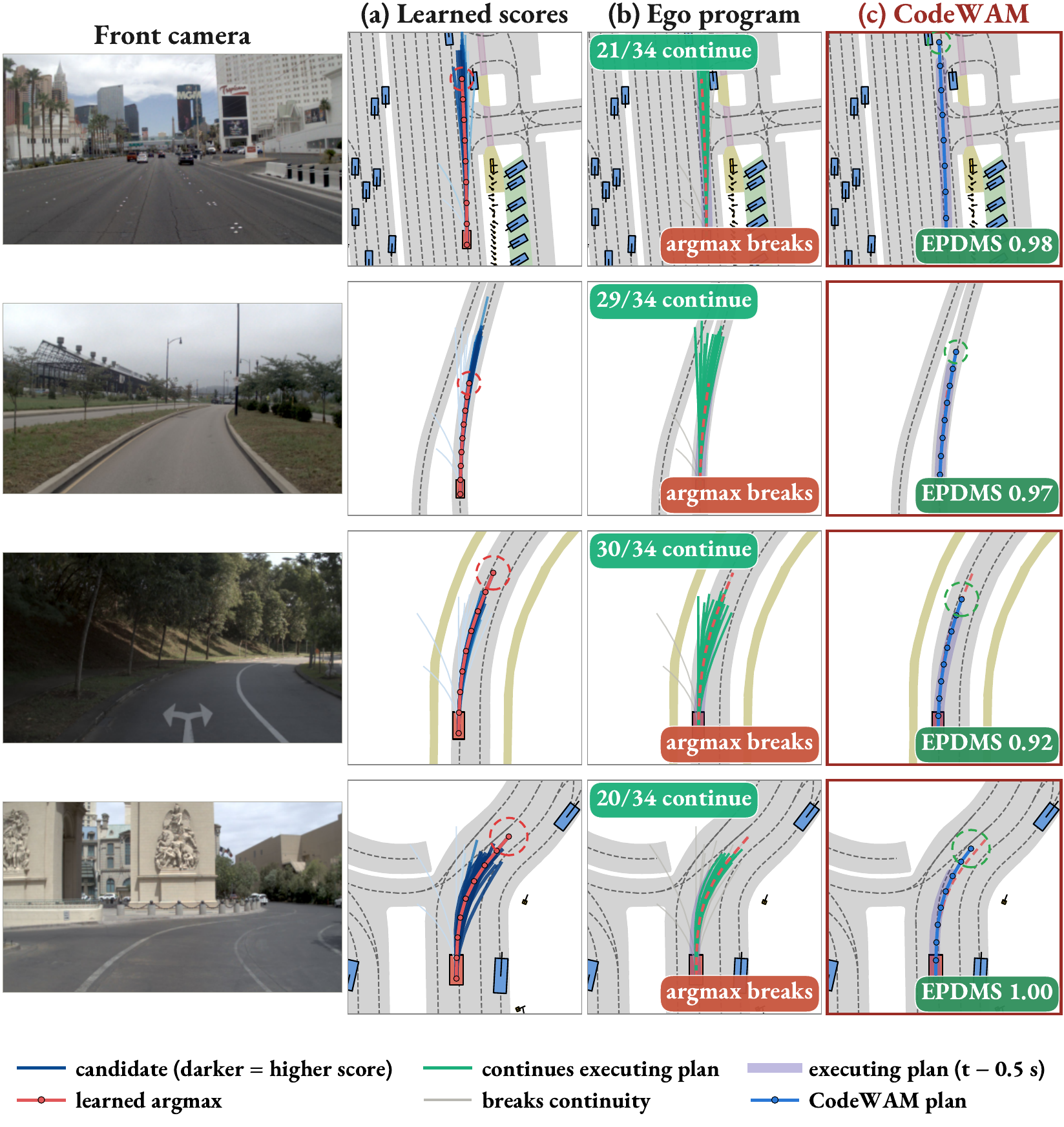}
\caption{Learned ranking and execution continuity in four scenes. In (a), darker blue indicates a higher learned candidate score and red marks the learned argmax. In (b), green and gray distinguish candidates that pass or fail the continuation test; counts give the number passing out of 34 candidates. The purple band denotes the plan issued 0.5\,s earlier. In (c), blue shows the final CodeWAM plan and badges report frame EPDMS on $[0,1]$. The learned argmax fails the continuation test in every displayed scene despite the presence of compatible alternatives.}
\label{fig:supp-candidates-main}
\end{figure*}

\begin{figure*}[!tp]
\centering
\includegraphics[width=\textwidth,height=0.84\textheight,keepaspectratio]{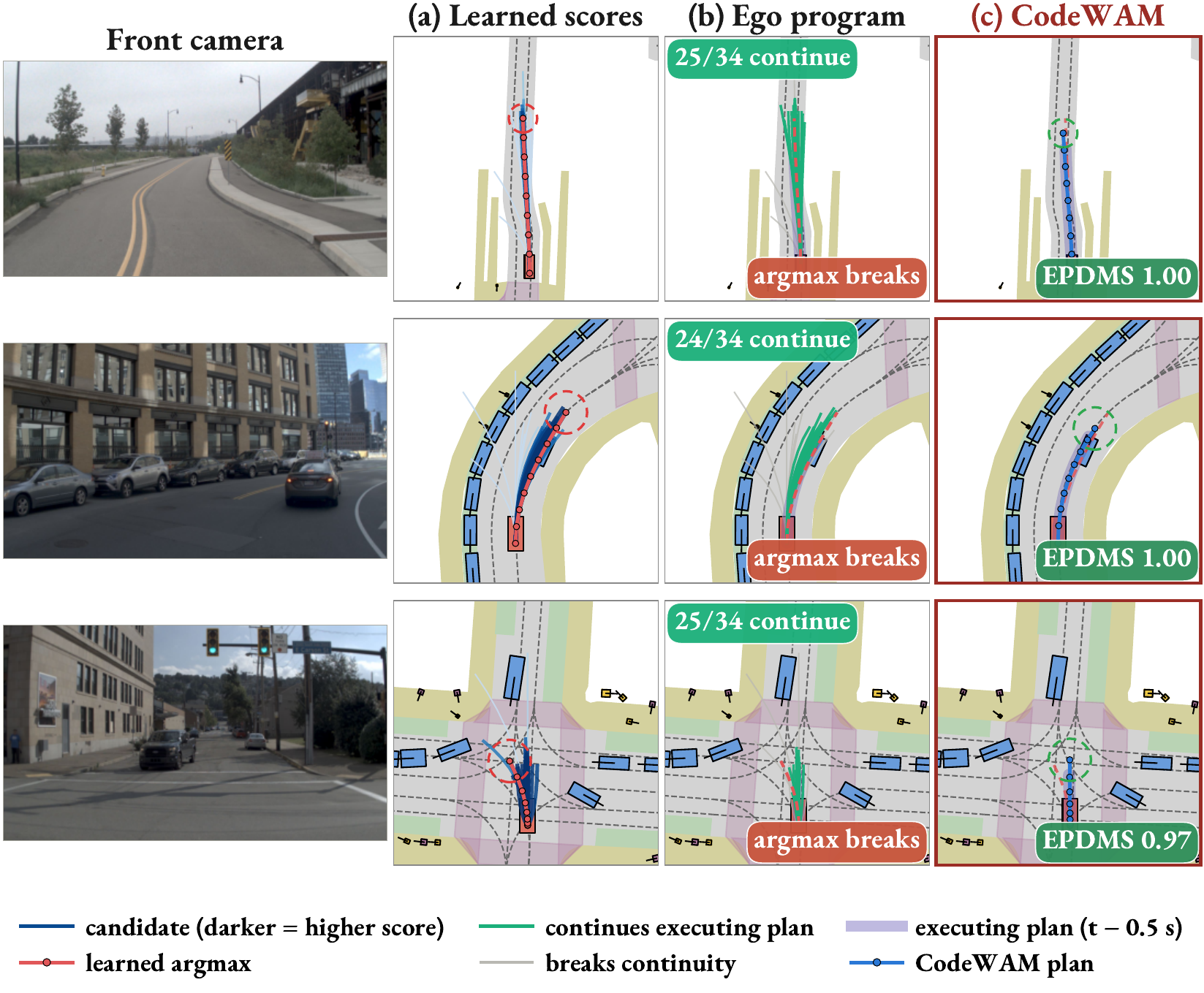}
\caption{Additional candidate-selection examples on a bend, a curved urban road, and an intersection approach, using the columns and colors of Figure~\ref{fig:supp-candidates-main}. From top to bottom, 25, 24, and 25 of the 34 candidates pass the continuation test, while the learned argmax fails it in each case. Final CodeWAM plans achieve frame EPDMS of 1.00, 1.00, and 0.97. Continuation counts measure compatibility with the preceding plan; EPDMS also accounts for scene-dependent consequences.}
\label{fig:supp-candidates-more}
\end{figure*}

Figures~\ref{fig:supp-candidates-main} and~\ref{fig:supp-candidates-more} compare the learned ranking with the program's assessment of continuity. Each scene contains 34 candidates. Between 20 and 30 pass the displayed continuation test, yet the learned argmax fails that test in every illustrated case. A high learned score therefore does not imply compatibility with the preceding plan. The purple band makes this distinction visible: candidates with similar spatial paths can differ in their transition from the motion already under execution. The final CodeWAM plan incorporates this execution context together with scene evaluation and refinement. Passing the continuation test alone does not certify safety or road compliance; these remain part of the joint consequence.

\subsection{Consistency across Consecutive Decisions}
\label{sec:supp-consecutive}

\begin{figure*}[!tp]
\centering
\includegraphics[width=\textwidth,height=0.84\textheight,keepaspectratio]{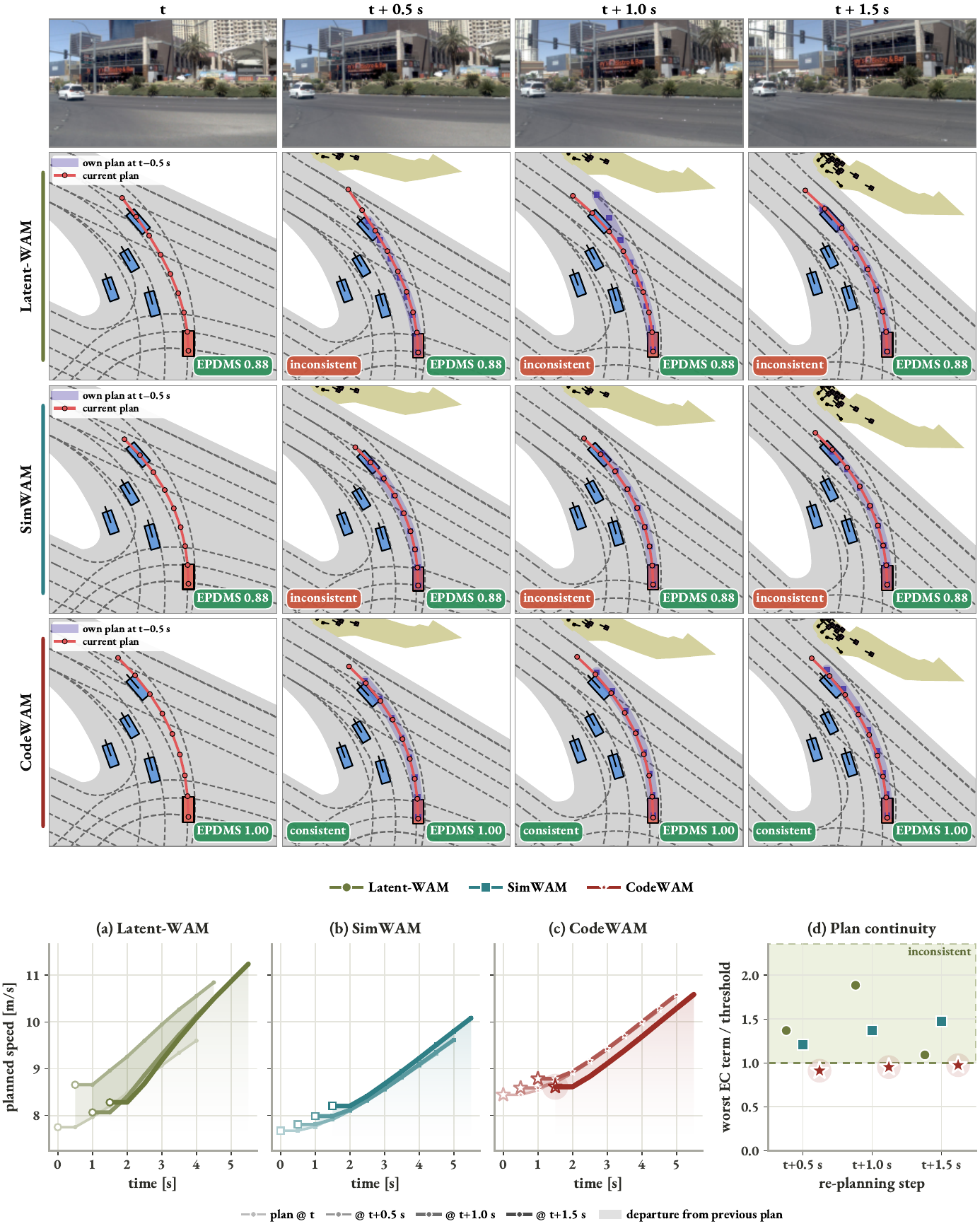}
\caption{Consecutive planning through a bend. Columns advance from $t$ to $t+1.5$\,s in 0.5-second steps; rows compare Latent-WAM, SimWAM, and CodeWAM. Red shows the current plan and purple the same method's preceding plan; badges report frame EPDMS. Lower panels (a--c) show planned speed profiles in m/s, with shading marking departures from the preceding plan. Panel (d) reports the largest threshold-normalized EC difference; values above one indicate inconsistency. CodeWAM remains below this threshold at all three updates, while both baselines exceed it.}
\label{fig:supp-strip-d}
\end{figure*}

\begin{figure*}[!tp]
\centering
\includegraphics[width=\textwidth,height=0.84\textheight,keepaspectratio]{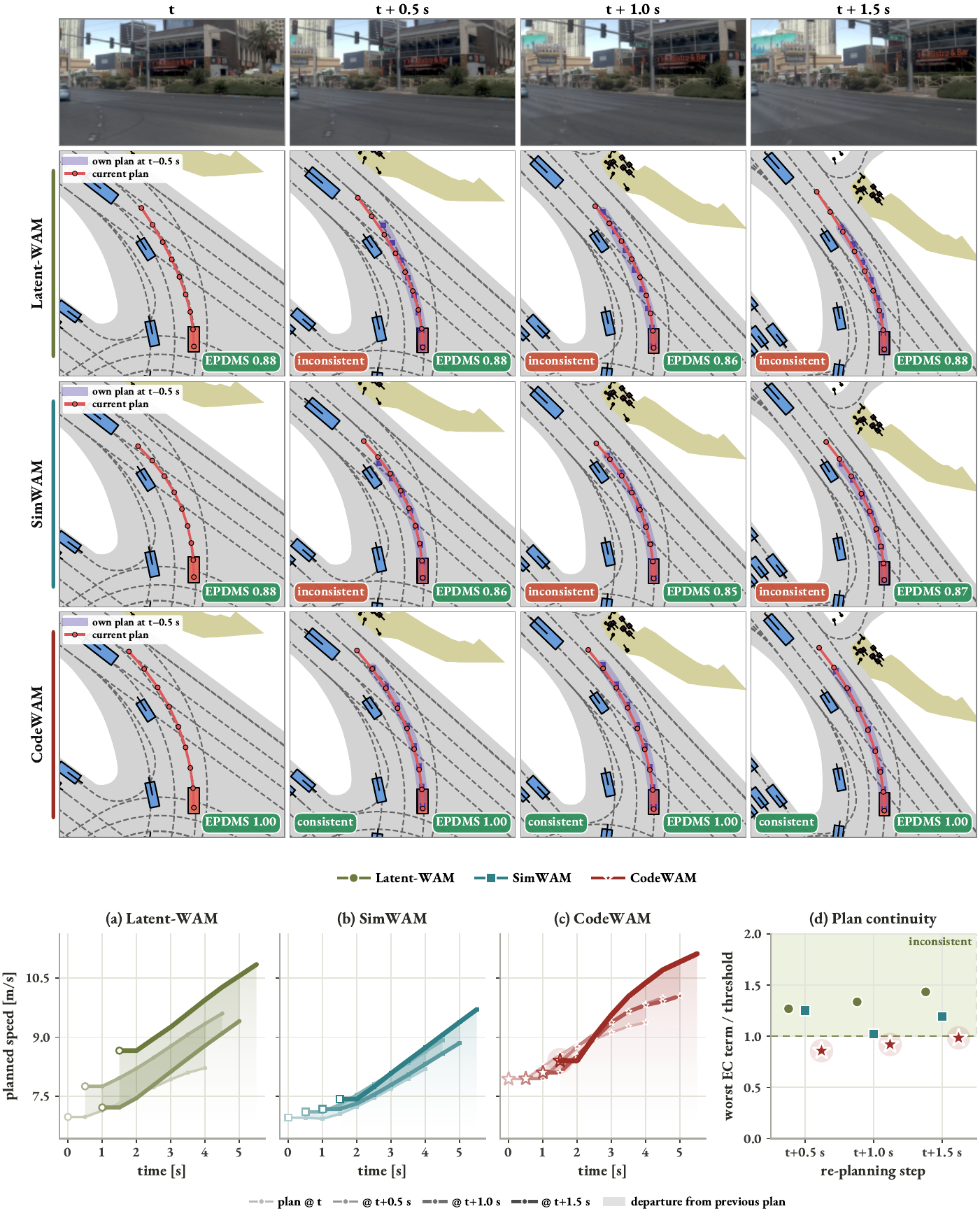}
\caption{Another planning window through the bend, using the layout and conventions of Figure~\ref{fig:supp-strip-d}. Both baselines fail the continuity test at all three updates. CodeWAM adjusts its planned speed while keeping the largest threshold-normalized EC difference below one and frame EPDMS at 1.00 throughout the window. The example shows that consistent decisions can accommodate changes in planned motion.}
\label{fig:supp-strip-e}
\end{figure*}

\begin{figure*}[!tp]
\centering
\includegraphics[width=\textwidth,height=0.84\textheight,keepaspectratio]{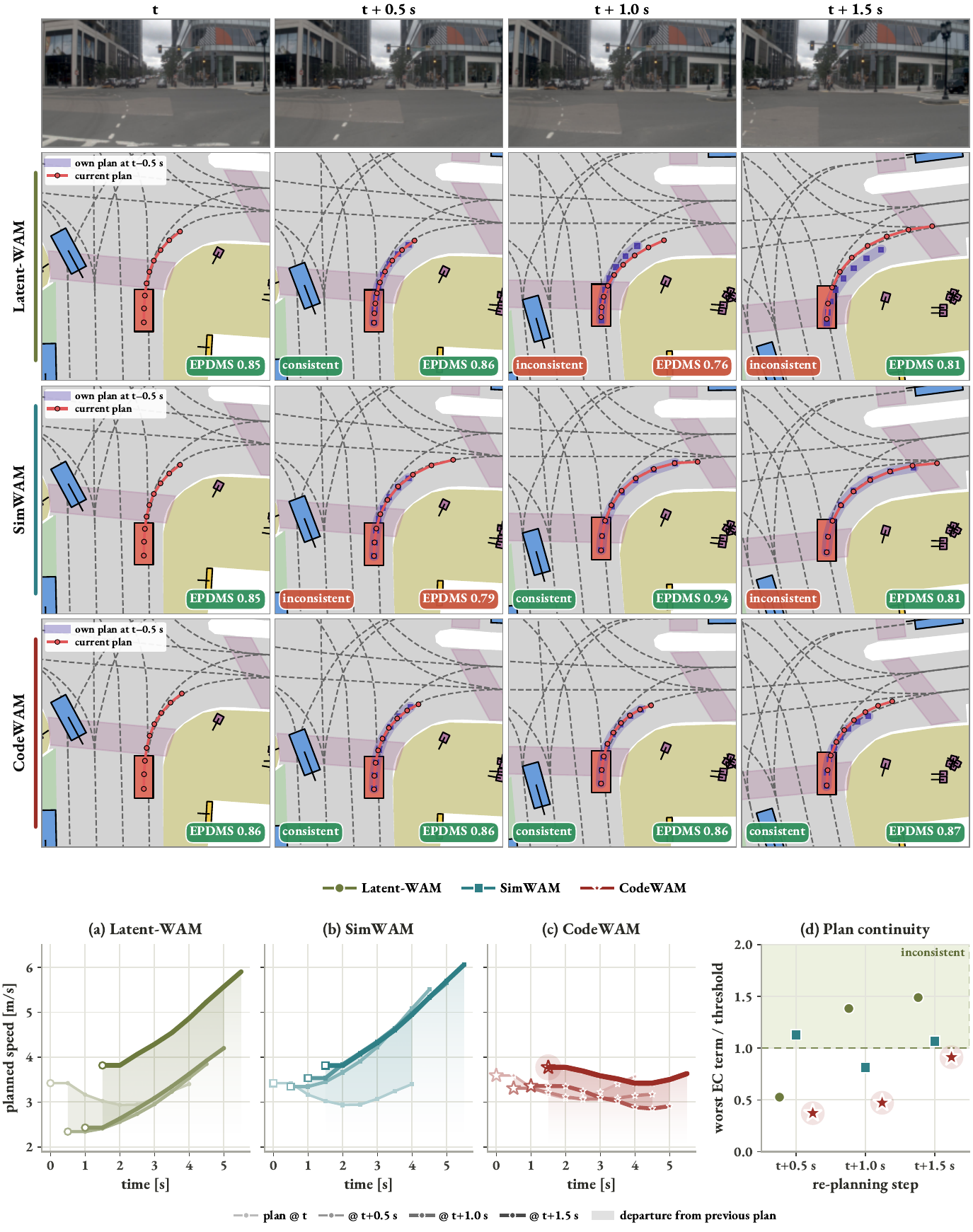}
\caption{Consecutive decisions during a right turn, using the layout and conventions of Figure~\ref{fig:supp-strip-d}. All three methods predict the same broad maneuver, but their planned speed profiles and cross-plan differences vary. CodeWAM passes all three continuity checks with frame EPDMS of 0.86--0.87. SimWAM reaches 0.94 at $t+1.0$\,s but fails consistency at $t+0.5$\,s and $t+1.5$\,s, illustrating that a higher individual-frame score need not imply consistent successive decisions.}
\label{fig:supp-strip-k}
\end{figure*}

Figures~\ref{fig:supp-strip-d}--\ref{fig:supp-strip-k} show four decisions at 0.5-second intervals. Each current plan is compared with the same planner's preceding plan, while the lower plots show the corresponding planned speed profiles. The continuity diagnostic normalizes each EC difference by its threshold and displays the largest ratio; values above one indicate a failed consistency test.

In the two bend windows, CodeWAM remains below this threshold at all three updates, whereas Latent-WAM and SimWAM exceed it. The right-turn example is more mixed: both baselines pass one update, but fail others, while CodeWAM passes all three. SimWAM nevertheless reaches a higher frame score at one instant (0.94 versus 0.86). This distinguishes temporal consistency from overall driving quality and supports evaluating both. The speed curves describe planned motion across updates, rather than measured acceleration or jerk from a physical closed-loop rollout.

\subsection{Planning from NAVHARD Second-Stage States}
\label{sec:supp-navhard-examples}

\begin{figure*}[!tp]
\centering
\includegraphics[width=\textwidth,height=0.84\textheight,keepaspectratio]{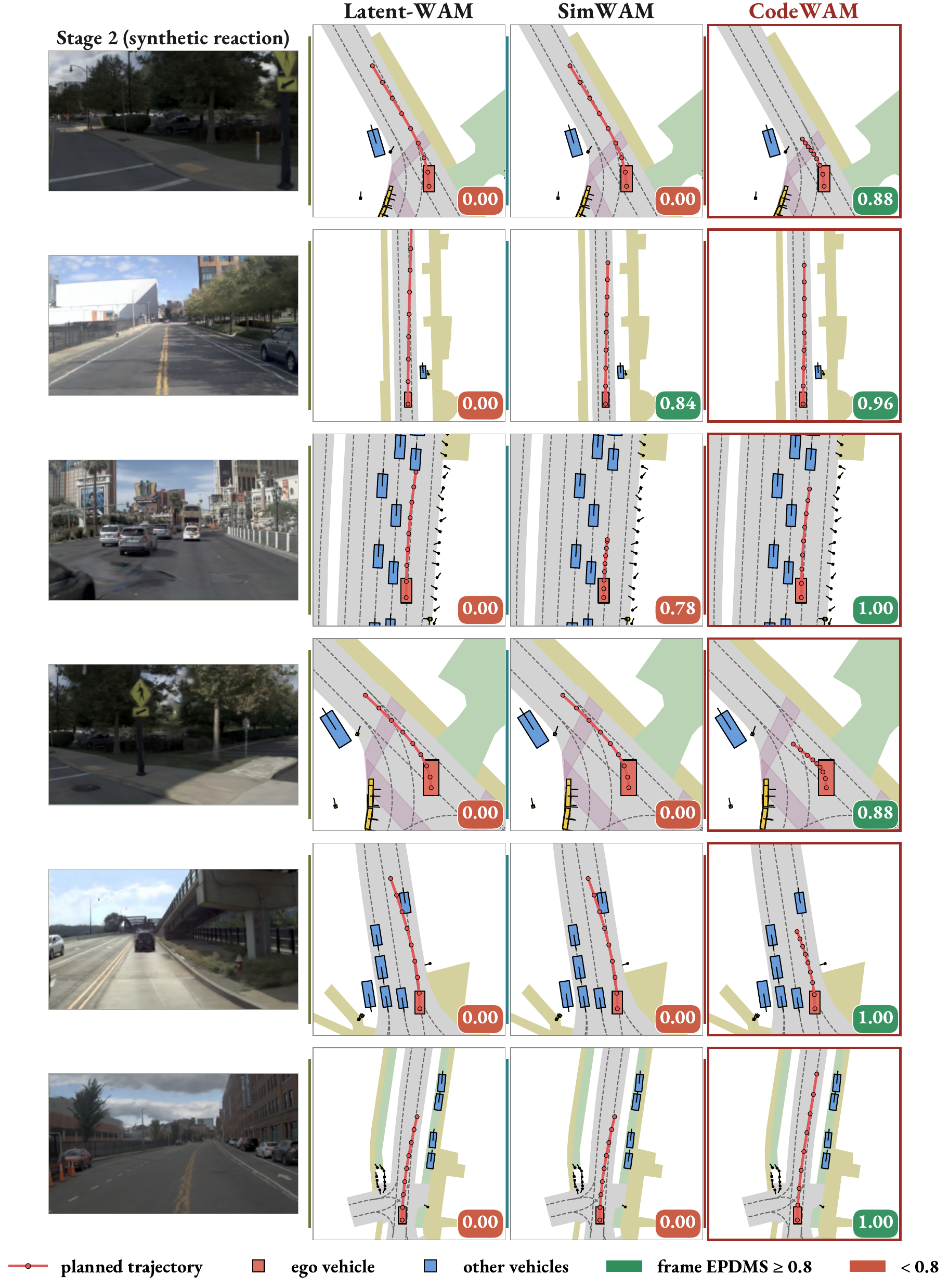}
\caption{Additional NAVHARD second-stage comparisons. Each row pairs a synthetic observation with plans from Latent-WAM, SimWAM, and CodeWAM. Red lines mark planned trajectories; salmon and blue boxes denote the ego vehicle and other vehicles. Badges report frame EPDMS on $[0,1]$, with green indicating $\geq 0.8$ and orange lower values. CodeWAM obtains nonzero scores in all six selected cases. These scores describe individual stage-2 plans, not the combined two-stage NAVHARD result.}
\label{fig:supp-navhard-more}
\end{figure*}

Figure~\ref{fig:supp-navhard-more} extends the two-stage examples in the main paper with six synthetic second-stage observations. These states test planning after the ego vehicle has departed from its recorded configuration. CodeWAM obtains frame scores from 0.875 to 1.000; Latent-WAM receives zero in these selected cases, while SimWAM retains nonzero scores in two. Some plans follow similar road geometry but differ in waypoint spacing and extent, illustrating that the consequence depends on how the motion unfolds as well as its spatial path. These frame-level examples complement, but do not replace, the proximity-weighted two-stage aggregation in Section~\ref{sec:supp-navhard}.

\subsection{Additional Zero-Shot Transfer to nuScenes}
\label{sec:supp-nusc-examples}

\begin{figure*}[!tp]
\centering
\includegraphics[width=\textwidth,height=0.84\textheight,keepaspectratio]{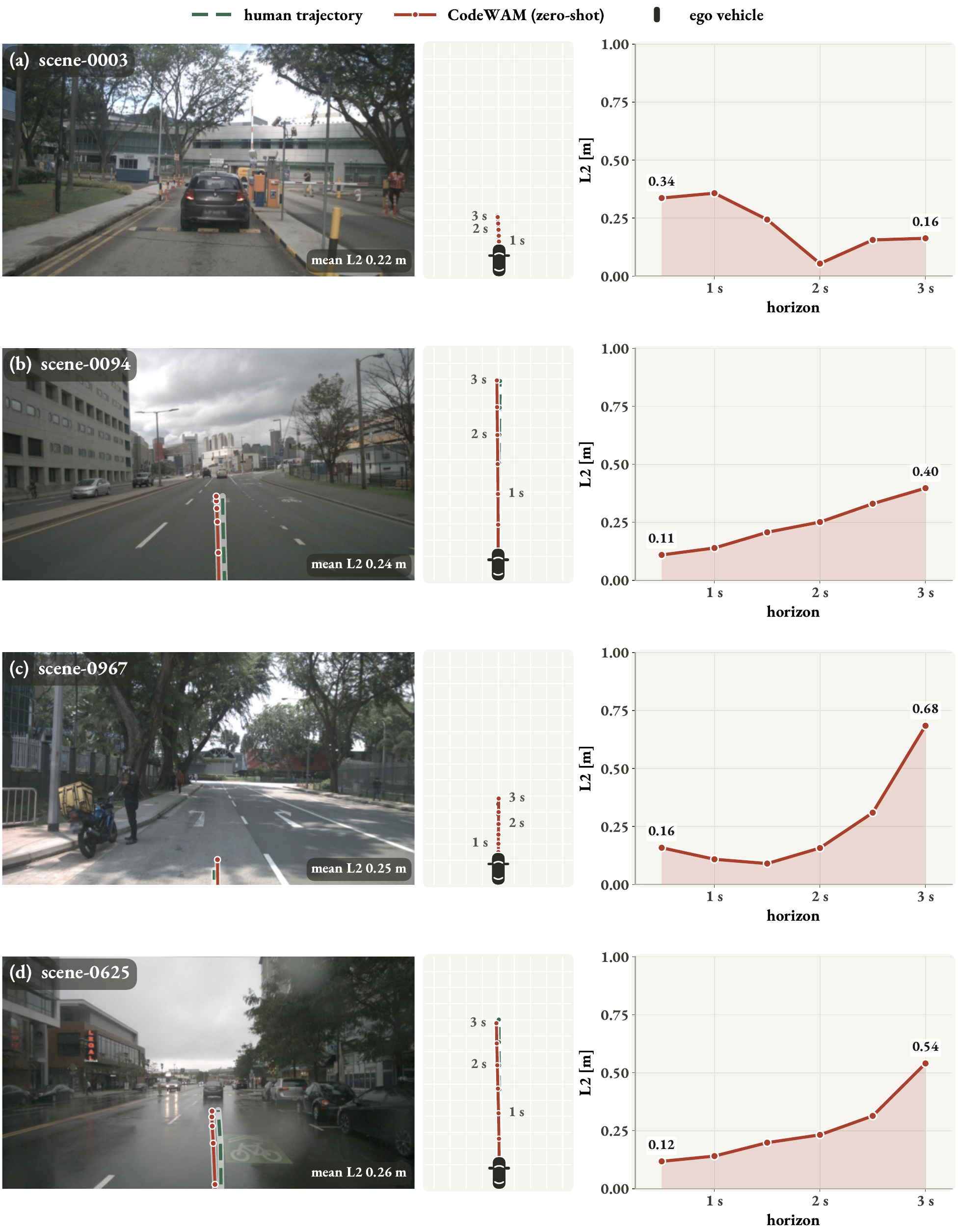}
\caption{Additional zero-shot predictions on nuScenes. Each row shows the camera observation, a bird's-eye comparison with recorded human motion, and pointwise L2 error in meters at 0.5-second intervals from 0.5 to 3\,s. Red denotes CodeWAM and dashed green the human trajectory. Image badges average the six errors within each example. The curves show both increasing long-horizon error and non-monotonic deviations; these per-example means differ from the average of the three horizon-averaged metrics in Table~\ref{tab:nuscenes}.}
\label{fig:supp-nusc-more}
\end{figure*}

Figure~\ref{fig:supp-nusc-more} shows four further nuScenes examples without fine-tuning. The predicted paths remain close to the recorded motion in the displayed road and lighting conditions, with six-sample mean L2 errors of 0.218, 0.239, 0.251, and 0.257\,m. The error curves also expose differences hidden by a single mean: scene-0003 improves after the early prediction steps, whereas scene-0967 has a larger discrepancy near 3\,s. Here each curve plots displacement error at the individual timestamp, and each image badge averages all six timestamps. This per-example average is distinct from averaging the three horizon-averaged entries in Table~\ref{tab:nuscenes}. The examples illustrate open-loop transfer rather than closed-loop safety.

\clearpage

\end{document}